\documentclass[lettersize,journal]{IEEEtran}
\usepackage{amsmath,amsfonts}
\usepackage{algorithmic}
\usepackage{algorithm}
\usepackage{microtype}
\usepackage{array}
\usepackage[caption=false,font=footnotesize,labelfont=sf,textfont=sf]{subfig}
\usepackage{textcomp}
\usepackage{stfloats}
\usepackage{url}
\usepackage{verbatim}
\usepackage{graphicx}
\usepackage{cite}
\usepackage{enumitem}
\usepackage{latexsym}
\usepackage{epstopdf}
\usepackage{amsthm}
\usepackage{upgreek}

\usepackage{booktabs}
\usepackage{lipsum}
\usepackage{tabularx}
\usepackage{multirow}
\usepackage{color}
\usepackage{threeparttable} 
\usepackage{amssymb}
\usepackage{makecell}
\usepackage{dsfont}
\usepackage[]{hyperref}  
\usepackage{float}
\usepackage[table]{xcolor}
\begin{document}
	
	\title{BMCR: Adaptive Backbone Module Composition via Reinforcement Learning for Remote Sensing Object Detection}          
	\author{Wenlin~Liu, Xikun~Hu,
		and~Ping~Zhong,~\IEEEmembership{Senior Member,~IEEE}
		% stops a space
		\thanks{Wenlin Liu, Xikun Hu, and Ping Zhong are with the College of Electronic Science and Technology, National University of Defense Technology, Changsha 410073, China (e-mail: wenlinliu@nudt.edu.cn; xikun@nudt.edu.cn; zhongping@nudt.edu.cn).}
		
		\thanks{This work was supported by the National Natural Science Foundation of China under Grant 62301574. (\textit{Corresponding authors: Xikun Hu and Ping Zhong})}
	}

	% The paper headers
	\markboth{Journal of \LaTeX\ Class Files,~Vol.~14, No.~8, August~2021}%
	{Shell \MakeLowercase{\textit{et al.}}: A Sample Article Using IEEEtran.cls for IEEE Journals}
	
	\IEEEpubid{0000--0000/00\$00.00~\copyright~2021 IEEE}
	% Remember, if you use this you must call \IEEEpubidadjcol in the second
	% column for its text to clear the IEEEpubid mark.
	
	\maketitle
	
	\begin{abstract}
		In remote sensing object detection, Convolutional Neural Networks (CNNs) excel at capturing local details while Vision Transformers (ViTs) are better at global context modeling. However, existing detectors typically rely on a single fixed backbone or a manually designed hybrid architecture, and thus fail to adaptively exploit these complementary strengths across inputs of diverse complexity. To address this limitation, we propose Backbone Module Composition via Reinforcement Learning (BMCR). BMCR dynamically assembles input-adaptive inference paths from reusable modules decomposed from off-the-shelf CNN and ViT backbones. To enable such cross-family composition, we first construct an extensible module toolbox. Specifically, we decompose representative CNN and ViT backbones into reusable functional modules and encapsulate each module with explicit structural, semantic, and computational metadata for compatibility-aware assembly. To bridge the gap between grid-based CNN features and token-based ViT representations, we design a lightweight Optimal Transport (OT) based transition interface that ensures distribution-aware alignment while respecting spatial consistency. The backbone composition process is then formulated as a sequential decision problem, in which a policy network progressively selects task-relevant modules according to intermediate multi-scale observations. To stabilize the joint optimization of reusable modules and the routing policy, we further develop an Adaptive Module Cooperative Optimization (AMCO) strategy that coordinates module updating, routing exploration, and reward assignment during training. On DOTA-v1.0, DOTA-v1.5 and DIOR-R, BMCR achieves 79.31\%, 73.41\% and 71.86\% mAP, respectively, surpassing strong static and dynamic baselines by up to 2.5 points while maintaining competitive efficiency.
	\end{abstract}

	
	\section{Introduction}
	\label{sec:intro}
	
	Recent advances in satellite imaging have produced remote sensing images with broader coverage and finer resolution~\cite{tuia2024artificial}. Unlike natural images, remote sensing scenes often exhibit strong spatial heterogeneity, where vast homogeneous regions coexist with dense small objects and complex man-made structures~\cite{cornebise2022open}. This poses a dual requirement for visual backbones: capturing fine-grained local details and long-range contextual dependencies while adapting their representational capacity to different input contents. Although high-capacity remote sensing foundation models have demonstrated strong representation ability~\cite{Guo_2024_CVPR,2023RingMo}, uniformly applying a large backbone to all inputs is computationally expensive and often redundant for simple scenes.
	
	At the architecture level, CNNs~\cite{he2016resnet} and ViTs~\cite{dosovitskiy2021an} provide complementary properties: CNNs are effective at modeling local spatial patterns, whereas ViTs are better suited for capturing global contextual dependencies~\cite{zhao2021battle, chen2025dynamicvis}. However, their relative importance varies significantly across remote sensing scenes. Most existing detectors instantiate these architectures as complete and fixed backbones, either as a single monolithic backbone or as a manually designed static hybrid architecture~\cite{hatamizadeh2023fastervit, yu2021path, wei2024pathnet}. Such input-independent designs impose the same computational path on all inputs, leading to redundant computation in simple scenes and limited adaptability in complex ones.

	\IEEEpubidadjcol
	Dynamic architectures have been explored to reduce the rigidity of fixed backbones by adapting computation to input contents~\cite{LI2022102926,hao2024efficient}. Existing methods usually adjust the execution of a predefined architecture, such as skipping layers, selecting branches, pruning tokens, or activating different subnetworks. Although these strategies improve computational flexibility, most of them operate under a homogeneous assumption: candidate operations belong to the same backbone family and share compatible feature formats. Consequently, the routing decision mainly determines how much of a fixed architecture to execute, rather than which heterogeneous architectural primitives to compose.
	
	This limitation is particularly restrictive for remote sensing object detection, where large-scale aerial images contain highly heterogeneous regions, ranging from homogeneous backgrounds to dense small-object clusters and complex man-made structures~\cite{cornebise2022open,Xia_2018_CVPR}. Such diversity makes no single architectural style uniformly optimal. However, reusing CNN and ViT modules as extensible architectural resources introduces mismatches in feature formats, spatial resolutions, channel dimensions, and semantic stages. As a result, existing dynamic networks cannot directly support heterogeneous, cross-family backbone composition, limiting their ability to exploit the complementary strengths of off-the-shelf architectures.

	\begin{figure}[t]
		\centering
		\includegraphics[width=1\columnwidth]{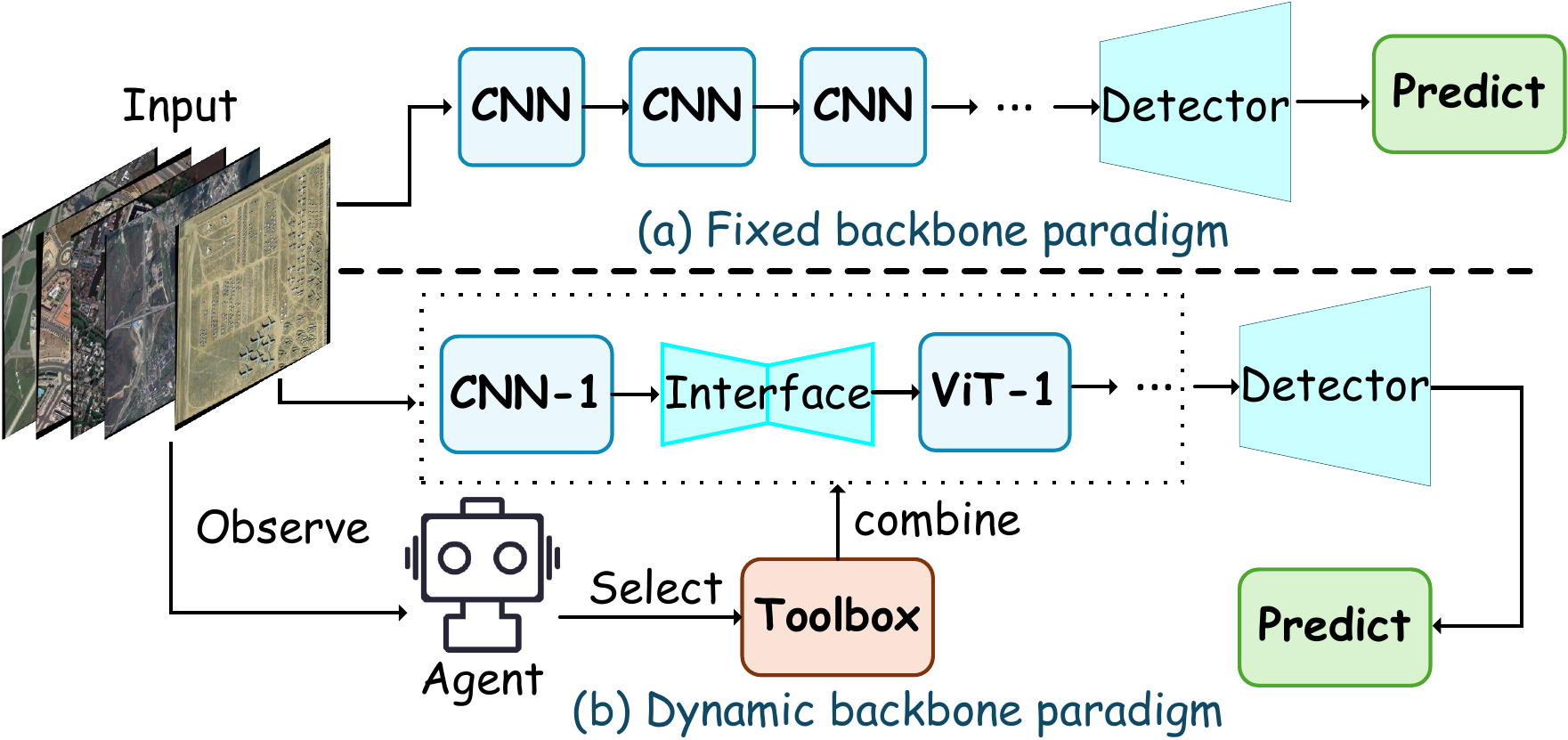}
		\caption{Comparison of backbone design paradigms. 
			(a) Fixed backbone: applies a static, uniform sequence of network layers to all inputs regardless of content. 
			(b) Dynamic backbone (BMCR): routes images through sample-adaptive execution paths composed of CNN and ViT modules on the fly.}
		\label{fig:simple}
	\end{figure}
	
	To address these limitations, we propose Backbone Module Composition
	via Reinforcement Learning (BMCR), an adaptive backbone composition
	framework for remote sensing object detection. Rather than designing
	another fixed backbone or searching within a homogeneous architecture,
	BMCR treats off-the-shelf CNN and ViT components as reusable modules
	and dynamically composes them into input-specific inference paths. This
	process involves discrete, sequential, and compatibility-constrained
	decisions, since each selected module changes the intermediate
	representation and determines the feasible choices and computational
	cost of subsequent steps. We therefore formulate backbone composition
	as a reinforcement learning problem, where a policy network
	progressively selects task-relevant modules according to intermediate
	observations. To support heterogeneous composition and stable training,
	BMCR further introduces an OT-based transition interface and an
	Adaptive Module Cooperative Optimization (AMCO) strategy. The main
	contributions of this work are as follows:
	
	\begin{itemize}
		\item We propose BMCR, an input-adaptive backbone composition framework for remote sensing object detection. Different from fixed or manually hybridized backbones, BMCR decomposes existing backbones into reusable modules and dynamically constructs sample-specific inference paths according to input contents.
		
		\item We build an extensible heterogeneous module toolbox with explicit structural, semantic, and computational metadata, and introduce an OT-based transition interface to support compatibility-aware composition between grid-based CNN features and token-based ViT representations across different spatial granularities.
		
		\item We formulate backbone composition as a sequential decision problem and develop AMCO to coordinate module updating, routing exploration, and reward assignment.
	\end{itemize}

	\section{Related Work}
	\label{sec:related_work}
	
	\subsection{Dynamic Inference and Neural Routing}
	
	Dynamic neural networks adapt inference to individual samples to improve the trade-off between task performance and computational cost. Existing methods adjust network depth through early exiting~\cite{teerapittayanon2016branchynet, ICLR2025_BEEM} or layer skipping~\cite{wang2018skipnet, fan2024not, luo2024skipdiff}, modulate local operators according to input content~\cite{chen2020dynamic, dai2017deformable, wang2023internimage, li2023lsknet}, or select input-dependent paths from predefined layers, branches, and operations~\cite{yu2021path, wei2024pathnet}. Although effective, these methods mainly adapt the execution of a predefined architecture, rather than composing new inference paths from independently developed backbone modules. Therefore, they provide limited support for reusing heterogeneous modules from off-the-shelf CNN and ViT backbones. This limitation is particularly relevant to remote sensing object detection, where different scenes may require different combinations of local spatial modeling, multi-scale representation, and global context reasoning. In contrast, BMCR targets cross-architecture backbone module composition and constructs input-adaptive paths from reusable components.
	
	\subsection{Heterogeneous Architecture Fusion}
	
	The complementary properties of CNNs and ViTs have motivated extensive studies on heterogeneous architecture fusion~\cite{lou2025transxnet}. Existing hybrid architectures combine convolutional and Transformer-based representations through fixed designs, such as parallel pathways~\cite{chenvision, peng2021conformer}, cascaded stacking~\cite{dai2021coatnet, hatamizadeh2023fastervit}, and interleaved block embedding~\cite{li2022next, meng2024cta}. Although effective, these methods are usually architecture-specific and rely on manually designed connectors between predetermined components. Such connectors are tailored to fixed hybrid backbones, rather than general transition interfaces for reusable modules from different pretrained architectures, limiting flexible module reuse and dynamic cross-architecture recombination. BMCR is designed to handle this issue with an OT-based interface for distribution-aware alignment between grids and tokens.
	
	\subsection{Reinforcement Learning for Adaptive Remote Sensing}
	
	Reinforcement learning (RL) provides a natural framework for adaptive
	decision-making with discrete, sequential, and task-dependent actions.
	In remote sensing, RL has been applied to sample
	selection~\cite{uzkent2020learning}, hyperspectral band
	selection~\cite{mou2021deep}, and content-aware receptive field
	adjustment~\cite{liu2023seeing, liu2024scale}. However, these methods
	operate at the data, spectral, or local operator level; backbone-level
	architectural routing remains underexplored. Unlike RL-based neural
	architecture search (NAS)~\cite{zheng2018learning}, which produces a
	fixed architecture after training, BMCR performs input-dependent
	routing at inference time. This setting is more challenging: each
	decision alters the intermediate representation and constrains
	subsequent feasible choices. BMCR addresses this by formulating
	backbone composition as a compatibility-constrained sequential decision
	process, where metadata-driven action masking restricts the agent to
	structurally valid modules at each step, keeping exploration tractable
	in the large heterogeneous action space. To jointly optimize the
	policy, the modules, and the detection head without divergence, BMCR
	introduces the AMCO strategy, which stabilizes training through
	random-path warm-up, auxiliary supervision, and progressive prior
	annealing.

	\section{Methodology}
	\label{sec:methodology}
	
	\begin{figure*}[t]
		\centering
		\includegraphics[width=0.98\textwidth]{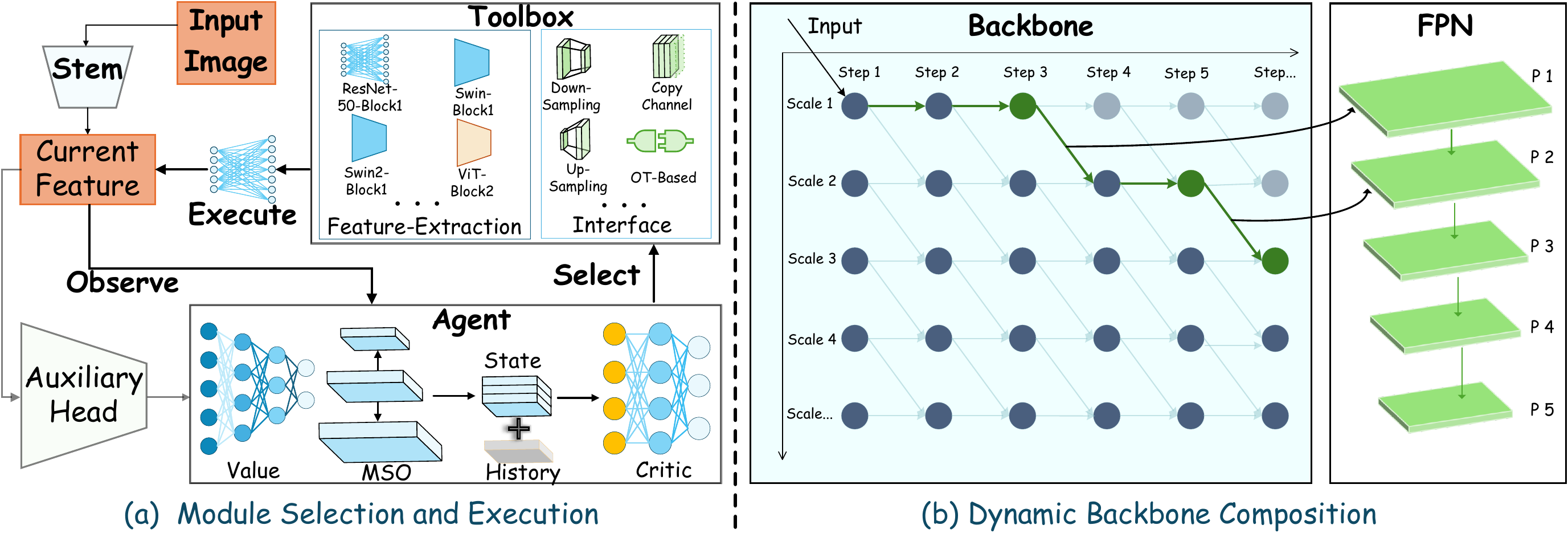}
		\caption{Overview of the proposed BMCR framework.
			(a) The routing agent observes intermediate features, selects task-relevant feature-extraction modules from the toolbox, and executes them with automatically determined transition interfaces.
			(b) BMCR constructs dynamic backbone paths across different feature granularities and feeds channel-aligned routed features into an FPN-based detection neck.}
		\label{fig:figure_main}
	\end{figure*}
	
	The objective of BMCR is to generate input-adaptive backbone representations for remote sensing object detection under highly variable scene conditions. Given an input image $\mathbf{I}$, a stem module first produces the initial feature $\mathbf{x}_0=S(\mathbf{I})$. BMCR then constructs the backbone as a sequential decision-making process, in which a routing agent selects feature-extraction modules from a heterogeneous toolbox based on the current feature state. As shown in Fig.~\ref{fig:figure_main}, the framework contains two key procedures: input-adaptive module composition and dynamic pyramid aggregation. The former builds a sample-specific backbone path, while the latter aligns and aggregates the routed multi-resolution features into a channel-aligned dynamic feature pyramid for FPN-based detection.
	
	The component selection and execution process in Fig.~\ref{fig:figure_main}(a) is formulated as an ``observe--select--execute'' Markov decision process (MDP). At the $t$-th step, the agent observes the state $\mathbf{s}_t=[\mathbf{x}_t,\mathbf{h}_t,\mathbf{p}_t]$, where $\mathbf{x}_t$ is the current intermediate feature, $\mathbf{h}_t$ records the routing history, and $\mathbf{p}_t$ is the task-specific spatial prior predicted by the auxiliary head. Based on $\mathbf{s}_t$, the policy network samples a valid action $a_t \sim \pi_{\theta}(\cdot \mid \mathbf{s}_t)$ from the masked action set $\mathcal{A}^{\mathrm{valid}}_t$. The action $a_t$ selects the next feature-extraction module, while the transition interface is deterministically selected by metadata rather than treated as an action. Specifically, $\mathcal{I}_t=\mathcal{I}_{\mathrm{red}}$ is used for same-representation transitions such as CNN-to-CNN or ViT-to-ViT, and $\mathcal{I}_t=\mathcal{I}_{\mathrm{OT}}$ is used for CNN-to-ViT or ViT-to-CNN transitions. The feature is then updated by $\mathbf{x}_{t+1}=f_{a_t}(\mathcal{I}_t(\mathbf{x}_t))$, where $f_{a_t}(\cdot)$ denotes the selected module. This closed-loop process constructs the backbone path $\mathbf{x}_0 \rightarrow a_0 \rightarrow \mathbf{x}_1 \rightarrow a_1 \rightarrow \cdots \rightarrow \mathbf{x}_T$ until the termination action $a_{\mathrm{end}}$ is selected or the number of executed steps reaches $T_{\max}$. Each input image therefore induces a sample-specific inference trajectory $\tau(\mathbf{I})=\{a_t\}_{t=0}^{T-1}$.
	
	After routing, the selected modules form a dynamic backbone path tailored to the input sample. As shown in Fig.~\ref{fig:figure_main}(b), the trajectory can be represented on a scale-step grid, where columns denote routing steps and rows denote feature granularity levels induced by spatial-size changes during routing. Unlike conventional static backbones with predefined downsampling stages, BMCR does not rely on fixed backbone stages or a predefined set of output levels. Instead, it records the routed features exported at different spatial granularities, denoted as $\mathcal{X}(\mathbf{I})=\{(\mathbf{x}_i,\eta_i)\}_{i=1}^{N_{\mathrm{r}}}$, where $N_{\mathrm{r}}$ is sample-dependent and $\eta_i$ contains the corresponding channel and representation metadata. Since the channel dimensions of routed features may differ, BMCR uses a metadata-indexed adapter bank to project each feature into a unified FPN width, i.e., $\tilde{\mathbf{x}}_i=\mathcal{A}_{\eta_i}(\mathbf{x}_i)$. The adapted features are sorted according to their spatial granularity and fed into an FPN-based neck, yielding $\{\mathbf{P}_i\}_{i=1}^{N_{\mathrm{r}}}$. The detection head applies shared prediction layers across the resulting pyramid levels and predicts oriented object instances as $\hat{\mathbf{Y}}=\mathrm{Head}_{\mathrm{det}}(\{\mathbf{P}_i\}_{i=1}^{N_{\mathrm{r}}})$. In this way, BMCR preserves a sample-adaptive backbone topology while producing channel-aligned multi-scale features for FPN-based detection.

	\subsection{Feature-Extraction Toolbox}
	
	The feature-extraction toolbox is constructed by decomposing representative backbone architectures into reusable functional modules. For hierarchical architectures, we extract core computational units from different stages that correspond to different feature granularities, such as residual blocks in ResNet~\cite{he2016resnet}. For isotropic architectures such as ViT, consecutive transformer blocks are grouped into reusable block-level primitives according to their depth positions. Each decomposed module preserves its original internal structure, including normalization layers, activation functions, and residual connections, so that its architectural inductive biases and feature transformation capacity are retained during composition.
	
	To provide a heterogeneous routing space, the toolbox includes six backbone-derived modules and two lightweight refinement modules. The backbone modules are derived from ResNet~\cite{he2016resnet}, ViT~\cite{dosovitskiy2021an}, Swin Transformer~\cite{liu2021swin}, SwinV2~\cite{swinv2}, ViTAE~\cite{xu2021vitae}, and ViTAE-RVSA~\cite{wang2022empirical}, while SELayer~\cite{hu2018senet} and CALayer~\cite{hou2021coordinate} provide cost-effective feature recalibration. Each module type is instantiated with three pretrained variants, which are registered as separate toolbox entries and optimized during BMCR training. This design expands representation diversity while keeping routing organized by module metadata.
	
	\subsection{Interface Toolbox}
	
	To support dynamic composition among heterogeneous modules, BMCR introduces an interface toolbox that is automatically invoked during routing. Given the current representation and the target module selected by the agent, the required interface is determined by their metadata rather than being selected as an additional action. The interface toolbox addresses two common compatibility issues: (i) \emph{structural mismatch}, caused by differences in spatial resolution and channel dimension between modules; and (ii) \emph{representational mismatch}, arising from the discrepancy between grid-based feature maps and sequence-based tokens.
	
	\subsubsection{Reduction Interface}
	
	The reduction interface handles structural mismatch between modules with the same representation format, such as CNN-to-CNN or ViT-to-ViT transitions. To keep routing efficient, it adopts parameter-free dimensional adaptation. Spatial mismatch is resolved by bilinear interpolation, and channel mismatch is handled by simple channel resampling through expansion or truncation. This provides a lightweight transition mechanism for non-adjacent or cross-stage modules. Since the reduction interface does not explicitly model cross-representation alignment, CNN--ViT transitions are delegated to the OT-based interface.
	
	\subsubsection{OT-Based Interface}
	
	The OT-based interface handles cross-representation transitions between grid-structured CNN feature maps and sequence-structured ViT tokens. Given the current representation and the target module, it first performs lightweight channel adaptation and then uses an optimal transport plan to remap source spatial locations to target tokens. A spatially regularized transport cost is used to preserve geometric consistency and discourage excessive long-range assignments. We describe the CNN-to-ViT case below, and the reverse ViT-to-CNN transition is obtained analogously.
	
	\paragraph{Distribution Modeling.}
	Given a CNN feature map $\mathbf{F}_s \in \mathbb{R}^{H_s \times W_s \times C_s}$, we flatten it into $\mathbf{X}_s \in \mathbb{R}^{M \times C_s}$ with $M=H_sW_s$, and apply a parameter-free channel adapter to match the target embedding dimension:
	\begin{equation}
		\widehat{\mathbf{X}}_s=\mathcal{A}_{\mathrm{ch}}(\mathbf{X}_s)\in \mathbb{R}^{M\times D_t}.
		\label{eq:channel_adaptation}
	\end{equation}
	Here, $D_t$ is the embedding dimension required by the target ViT module. For the target module, we define module-specific token anchors $\mathbf{E}_t=[\mathbf{e}_1^t,\mathbf{e}_2^t,\ldots,\mathbf{e}_N^t]^\top \in \mathbb{R}^{N\times D_t}$, where $N$ is determined by the target token budget. In our implementation, these anchors are learnable prototypes initialized from positional embeddings. Each source location and target token is assigned a normalized coordinate, denoted by $\mathbf{p}_i^s$ and $\mathbf{p}_j^t$, respectively. The source and target are then modeled as empirical measures:
	\begin{equation}
		\mu_s=\sum_{i=1}^{M}w_i^s\delta_{(\widehat{\mathbf{x}}_i^s,\mathbf{p}_i^s)},\quad
		\mu_t=\sum_{j=1}^{N}w_j^t\delta_{(\mathbf{e}_j^t,\mathbf{p}_j^t)},
		\label{eq:ot_measures}
	\end{equation}
	where uniform weights $w_i^s=1/M$ and $w_j^t=1/N$ are used.
	
	\paragraph{Cost Matrix and Transport Plan.}
	The transport cost between source location $i$ and target token $j$ combines feature dissimilarity and spatial displacement:
	\begin{equation}
		C_{ij}
		=
		\alpha
		\left(
		1-
		\frac{\widehat{\mathbf{x}}_i^s\cdot\mathbf{e}_j^t}
		{\|\widehat{\mathbf{x}}_i^s\|_2\|\mathbf{e}_j^t\|_2+\eta}
		\right)
		+
		(1-\alpha)
		\frac{\|\mathbf{p}_i^s-\mathbf{p}_j^t\|_2^2}{2\sigma^2},
		\label{eq:cost_matrix}
	\end{equation}
	where $\alpha$ balances semantic and spatial terms, $\sigma$ controls the spatial constraint, and $\eta$ ensures numerical stability. The entropy-regularized OT plan is obtained by
	\begin{equation}
		\mathbf{T}^*
		=
		\arg\min_{\mathbf{T}\in\mathcal{U}(\mathbf{w}^s,\mathbf{w}^t)}
		\langle \mathbf{T},\mathbf{C} \rangle
		+
		\varepsilon
		\sum_{i=1}^{M}\sum_{j=1}^{N}
		T_{ij}(\log T_{ij}-1),
		\label{eq:ot_plan}
	\end{equation}
	where $\mathcal{U}(\mathbf{w}^s,\mathbf{w}^t)$ denotes the transport polytope. This is solved using a fixed number of Sinkhorn iterations, and the target token budget is bounded to maintain a tractable cost matrix.
	
	\paragraph{Feature Remapping.}
	The adapted CNN features are remapped to target ViT tokens using barycentric projection:
	\begin{equation}
		\mathbf{X}_{s\rightarrow t}
		=
		\mathrm{diag}(\mathbf{w}^t)^{-1}
		(\mathbf{T}^*)^\top
		\widehat{\mathbf{X}}_s
		\in \mathbb{R}^{N\times D_t}.
		\label{eq:remap_features}
	\end{equation}
	The resulting $\mathbf{X}_{s\rightarrow t}$ matches the token length and embedding dimension required by the target ViT module. The reverse ViT-to-CNN transition follows the same formulation by treating spatial ViT tokens as the source and CNN grid locations as the target, followed by reshaping the remapped sequence into a grid feature map.
	
	\subsection{State Space}
	
	At the $t$-th routing step, the agent observes a compact state representation composed of the current intermediate feature, routing history, and spatial objectness prior. To handle heterogeneous routed features, the current feature $\mathcal{F}_t$ is resized to a common spatial resolution and projected to a unified channel dimension before state encoding.

	To capture spatial context at multiple granularities, we construct a Multi-Scale Pyramid (MSP) observation. The adapted feature is resized to several predefined resolutions $\mathcal{R}$, processed by lightweight convolutional blocks, and then resized back to a common spatial size $H\times W$ for concatenation:
	\begin{equation}
		\mathcal{F}_{t}^{\mathrm{cat}}
		=
		\mathrm{Concat}_{r\in\mathcal{R}}
		\left[
		\mathrm{Resize}_{H,W}
		\left(
		\mathrm{Conv}_{r}
		\left(
		\mathrm{Resize}_{r}(\mathcal{F}_t)
		\right)
		\right)
		\right].
		\label{eq:msp_concat}
	\end{equation}
	Here, $\mathrm{Concat}_{r\in\mathcal{R}}[\cdot]$ denotes concatenation over all predefined resolutions.
	The concatenated feature is projected into a unified state dimension:
	\begin{equation}
		\mathcal{F}_{t}^{\mathrm{MSP}}
		=
		\mathrm{Conv}_{1\times1}
		\left(
		\mathcal{F}_{t}^{\mathrm{cat}}
		\right)
		\in \mathbb{R}^{H \times W \times C_{\mathrm{state}}},
		\label{eq:msp_processing}
	\end{equation}
	where $C_{\mathrm{state}}$ is the state feature dimension. The predefined resolutions in $\mathcal{R}$ are used only for state observation and do not constrain the dynamic pyramid outputs.
	
	To encode routing history, the most recent $D$ actions are embedded into a history vector $\mathbf{h}_t=\mathrm{Embed}(a_{t-D},\ldots,a_{t-1})$ and broadcast to the state resolution:
	\begin{equation}
		\mathcal{F}_{t}^{\mathrm{hist}}
		=
		\mathrm{Broadcast}_{H,W}(\mathbf{h}_t)
		\in \mathbb{R}^{H \times W \times C_h}.
		\label{eq:history_map}
	\end{equation}
	
	Missing entries at early routing steps are zero-padded. In addition, an auxiliary objectness head predicts a spatial prior from the current feature:
	
	\begin{equation}
		\mathcal{F}_{t}^{\mathrm{obj}}
		=
		\mathrm{Resize}_{H,W}
		\left(
		\sigma(g_{\mathrm{aux}}(\mathcal{F}_t))
		\right)
		\in \mathbb{R}^{H \times W \times 1}.
		\label{eq:spatial_prior}
	\end{equation}
	Ground-truth box masks supervise only the auxiliary head and are not fed into the routing policy. The predicted prior is gradually annealed in the agent state.

	The final state observation is obtained by concatenation:
	\begin{equation}
		\mathbf{s}_t
		=
		\mathrm{Concat}
		\left(
		\mathcal{F}_{t}^{\mathrm{MSP}},
		\mathcal{F}_{t}^{\mathrm{hist}},
		\mathcal{F}_{t}^{\mathrm{obj}}
		\right).
		\label{eq:state_space}
	\end{equation}
	After global pooling, the state observation is fed into lightweight policy and value heads for action prediction and value estimation.
	
	\subsection{Action Space}
	
	The action space consists of $K$ candidate feature-extraction modules and one termination action:
	\begin{equation}
		\mathcal{A}=\{a_1,a_2,\ldots,a_K,a_{\mathrm{end}}\}.
		\label{eq:action_space}
	\end{equation}
	Here, $a_k$ selects the $k$-th feature-extraction module, and $a_{\mathrm{end}}$ terminates the current path construction. The transition interface is not included in the action space. Once a target module is selected, BMCR automatically determines the required interface according to the source and target metadata: the reduction interface is used for transitions within the same representation type, whereas the OT-based interface is used for CNN--ViT transitions. At each step, structurally invalid actions are masked out, and the agent samples from the masked policy:
	\begin{equation}
		a_t \sim \pi_{\theta}^{\mathrm{mask}}(\cdot \mid \mathbf{s}_t),
		\quad a_t\in \mathcal{A}^{\mathrm{valid}}_t .
		\label{eq:policy_sampling}
	\end{equation}
	This prevents the agent from selecting modules that violate transition constraints or routing validity.

	\subsection{Reward Function}
	
	The reward function is designed to encourage feature-quality improvement while penalizing high computation cost, redundant routing, and frequent CNN--ViT switches. At each routing step, the intermediate feature is fed into the auxiliary detection head to estimate a normalized feature-quality score:
	\begin{equation}
		q_t = Q_{\mathrm{aux}}(\mathcal{F}_t),
		\label{eq:feature_quality}
	\end{equation}
	where $Q_{\mathrm{aux}}(\cdot)$ maps the auxiliary predictions of $\mathcal{F}_t$ to a scalar detection-quality score. For a non-termination action, the quality improvement is defined as
	\begin{equation}
		\Delta q_t = q_{t+1}-q_t .
		\label{eq:quality_improvement}
	\end{equation}
	When $a_t=a_{\mathrm{end}}$, no feature transformation is executed and $\Delta q_t$ is set to zero.
	
	The computation cost includes the FLOPs of the selected module and the required transition interface:
	\begin{equation}
		c_t
		=
		\frac{
			\mathrm{FLOPs}(f_{a_t})
			+
			\mathrm{FLOPs}_{\mathrm{int}}(t,a_t)
		}{
			C_{\mathrm{ref}}
		},
		\quad a_t \neq a_{\mathrm{end}} .
		\label{eq:cost_term}
	\end{equation}
	For the termination action, we set $c_t=0$. Here, $f_{a_t}$ denotes the module selected by action $a_t$, $\mathrm{FLOPs}_{\mathrm{int}}(t,a_t)$ denotes the cost of the transition interface required at step $t$, and $C_{\mathrm{ref}}$ is the FLOPs of a reference static backbone.
	
	To reduce unnecessary use of the OT interface, we penalize cross-family transitions as
	\begin{equation}
		S_t
		=
		\mathbf{1}[a_t \neq a_{\mathrm{end}}]
		\mathbf{1}[\mathrm{OT}(t,a_t)=1],
		\label{eq:switch_penalty}
	\end{equation}
	where $\mathrm{OT}(t,a_t)=1$ indicates that the transition required by action $a_t$ at step $t$ invokes the OT-based interface.
	
	The termination reward is computed from the detection result obtained after the current path is terminated. Let
	\begin{equation}
		d_t
		=
		\mathrm{mAP}
		\left(
		\hat{\mathbf{Y}}_t,\mathbf{Y}
		\right),
		\label{eq:terminal_score}
	\end{equation}
	where $\hat{\mathbf{Y}}_t$ is the detection output of the terminated path $\mathcal{P}_t$, $\mathbf{Y}$ denotes the ground-truth annotations, and $\mathrm{mAP}(\cdot)$ is computed on the current sample and detached as a scalar terminal reward.
	\begin{equation}
		E_t
		=
		\mathbf{1}[a_t=a_{\mathrm{end}}]
		\mathbf{1}[\mathrm{Valid}(\mathcal{P}_t)]
		d_t .
		\label{eq:termination_reward}
	\end{equation}
	Here, $\mathrm{Valid}(\mathcal{P}_t)$ ensures that the terminated path satisfies the transition constraints and produces routed features that can be processed by the FPN-based neck.
	
	The immediate reward is finally formulated as
	\begin{equation}
		r_t
		=
		\lambda_{\mathrm{acc}}\Delta q_t
		-
		\lambda_{\mathrm{cost}}c_t
		-
		\lambda_{\mathrm{sw}}S_t
		+
		\lambda_{\mathrm{end}}E_t .
		\label{eq:reward_function}
	\end{equation}

	\subsection{Adaptive Module Cooperative Optimization Algorithm}
	\label{subsec:amco}
	
\begin{figure}[t]
		\centering
		\includegraphics[width=\columnwidth]{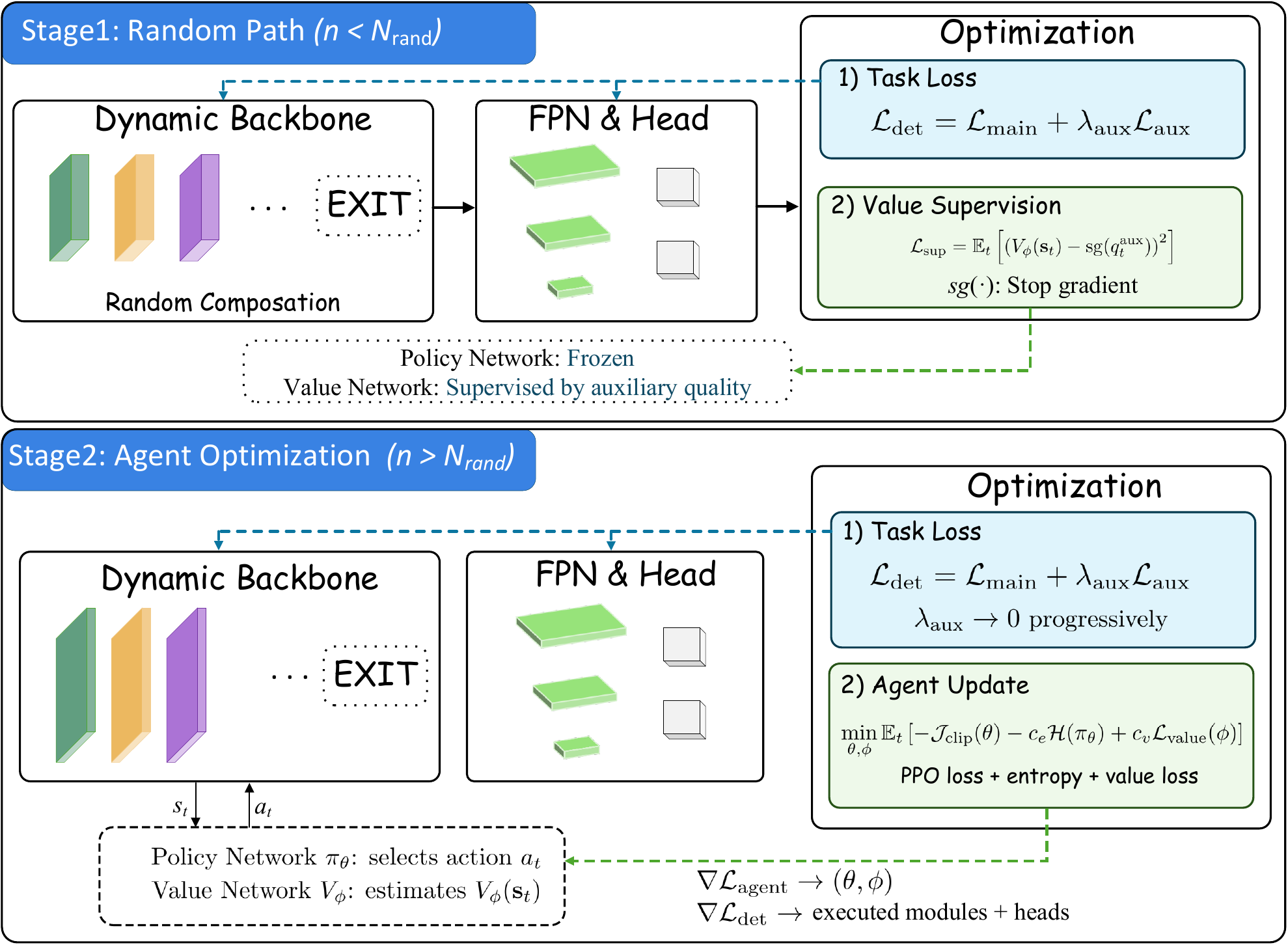}
		\caption{
			Illustration of the proposed AMCO algorithm. Stage~1 warms up the toolbox with random valid paths, while Stage~2 performs PPO-based routing optimization with decoupled agent and task updates.
		}
		
		\label{fig:amco}
	\end{figure}

	The proposed AMCO algorithm is designed to stabilize the joint training of the module combination agent, heterogeneous toolbox, and detection heads. Although the routing policy is optimized with PPO~\cite{schulman2017proximal}, directly applying PPO to BMCR is challenging because the agent operates in a large discrete module space, and detector-level feedback is delayed and sparse for step-wise routing optimization. To address these issues, AMCO introduces a two-phase cooperative optimization scheme that combines random-path warm-up, auxiliary feature-quality supervision, progressive objectness-prior annealing, and PPO-based policy refinement. The overall training procedure is shown in Fig.~\ref{fig:amco}.
	
	\subsubsection{Random-Path Warm-up Phase ($n \leq N_{\mathrm{rand}}$)}
	
	During early training, the routing policy is not updated, and actions are uniformly sampled from the structurally valid action set:
	\begin{equation}
		a_t \sim \mathrm{Uniform}
		\left(
		\mathcal{A}_{\mathrm{valid}}(\mathbf{s}_t)
		\right),
		\label{eq:random_policy}
	\end{equation}
	where $\mathcal{A}_{\mathrm{valid}}(\mathbf{s}_t)$ denotes the masked valid action set. The termination action is disabled until the current path satisfies the minimum requirements for FPN-based detection. This warm-up exposes the toolbox modules and detection heads to diverse valid paths and reduces early routing bias.
	
	During this phase, the executed modules, invoked transition interfaces, FPN neck, and main detection head are optimized by the detection loss $\mathcal{L}_{\mathrm{det}}$. The auxiliary head is trained with auxiliary objectness and quality targets to estimate the dense feature-quality score $q_t$ defined in Eq.~\eqref{eq:feature_quality}, which provides step-wise supervision for value learning.
	
	The value network is supervised by the auxiliary quality score:
	\begin{equation}
		\mathcal{L}_{\mathrm{sup}}
		=
		\mathbb{E}_{t}
		\left[
		\left(
		V_{\phi}(\mathbf{s}_t)
		-
		\mathrm{sg}(q_t)
		\right)^2
		\right],
		\label{eq:sup_loss}
	\end{equation}
	where $V_{\phi}(\cdot)$ is the value network and $\mathrm{sg}(\cdot)$ denotes stop-gradient. This supervision provides dense value targets before reliable policy gradients are available.
	
	The auxiliary head also predicts the objectness prior map $\mathcal{F}_{t}^{\mathrm{obj}}$ used in the agent state. Ground-truth box masks supervise only the auxiliary head and are not directly fed into the routing policy. During PPO training, the predicted prior is gradually annealed in the agent state.

	\subsubsection{PPO Routing Optimization Phase ($n > N_{\mathrm{rand}}$)}
	
	After random-path warm-up, AMCO switches to PPO-based routing optimization while continuing task-level training of the executed modules and detection heads. At each iteration, trajectories are collected using the old masked policy $\pi_{\theta_{\mathrm{old}}}^{\mathrm{mask}}$ and the current toolbox parameters. The collected state features are detached when updating the routing agent, so that PPO gradients only optimize the policy and value networks rather than directly modifying the feature-extraction modules.
	
	To bridge auxiliary-guided value learning and reward-driven policy optimization, we introduce a transition coefficient
	
	\begin{equation}
		\lambda_{\mathrm{tr}}(n)
		=
		\max
		\left(
		0,
		1-\frac{n-N_{\mathrm{rand}}}{N_{\mathrm{tr}}}
		\right),
		\label{eq:lambda_transition}
	\end{equation}
	where $N_{\mathrm{tr}}$ controls the transition duration. When $n$ is close to $N_{\mathrm{rand}}$, the value network relies more on the auxiliary quality target; as training proceeds, the value target gradually shifts toward RL returns.
	
	The auxiliary objectness prior is also annealed during PPO training:
	\begin{equation}
		\mu(n)
		=
		\max
		\left(
		0,
		1-\frac{n-N_{\mathrm{rand}}}{N_{\mathrm{prior}}}
		\right),
		\label{eq:prior_decay}
	\end{equation}
	where $N_{\mathrm{prior}}$ controls the decay duration. The state input is constructed as
	\begin{equation}
		\mathbf{s}_t^{(n)}
		=
		\mathrm{Concat}
		\left(
		\mathcal{F}_{t}^{\mathrm{MSP}},
		\mathcal{F}_{t}^{\mathrm{hist}},
		\mu(n)\mathcal{F}_{t}^{\mathrm{obj}}
		\right).
		\label{eq:state_with_prior_decay}
	\end{equation}
	For notational simplicity, the superscript $(n)$ is omitted below.
	
	Given the collected trajectory, the temporal-difference residual and generalized advantage estimate are computed as
	\begin{equation}
		\delta_t
		=
		r_t
		+
		\gamma V_{\phi}(\mathbf{s}_{t+1})
		-
		V_{\phi}(\mathbf{s}_t),
		\label{eq:td_residual}
	\end{equation}
	\begin{equation}
		\hat{A}_t
		=
		\sum_{l=0}^{T-t-1}
		(\gamma\lambda_{\mathrm{GAE}})^l
		\delta_{t+l},
		\label{eq:gae}
	\end{equation}
	where $\gamma$ is the discount factor, $\lambda_{\mathrm{GAE}}$ controls the bias--variance trade-off, and the next-state value is set to zero at the terminal step. The return target is
	\begin{equation}
		\hat{R}_t
		=
		\hat{A}_t
		+
		V_{\phi}(\mathbf{s}_t).
		\label{eq:return_target}
	\end{equation}

	The policy is optimized using the clipped PPO objective:
	\begin{equation}
		\mathcal{L}_{\mathrm{policy}}
		=
		-\mathbb{E}_{t}
		\left[
		\min
		\left(
		\rho_t\hat{A}_t,
		\bar{\rho}_t\hat{A}_t
		\right)
		\right],
		\label{eq:ppo_policy_loss}
	\end{equation}
	where $\rho_t(\theta)=\pi_{\theta}^{\mathrm{mask}}(a_t|\mathbf{s}_t)/
	\pi_{\theta_{\mathrm{old}}}^{\mathrm{mask}}(a_t|\mathbf{s}_t)$, 
	$\bar{\rho}_t=\mathrm{clip}(\rho_t,1-\epsilon,1+\epsilon)$, and $\epsilon$ is the clipping threshold.

	The value target combines the auxiliary quality score and the RL return:
	\begin{equation}
		y_t
		=
		\lambda_{\mathrm{tr}}(n)\mathrm{sg}(q_t)
		+
		\left(1-\lambda_{\mathrm{tr}}(n)\right)
		\mathrm{sg}(\hat{R}_t),
		\label{eq:value_target}
	\end{equation}
	\begin{equation}
		\mathcal{L}_{\mathrm{value}}
		=
		\mathbb{E}_{t}
		\left[
		\left(
		V_{\phi}(\mathbf{s}_t)-y_t
		\right)^2
		\right],
		\label{eq:value_loss}
	\end{equation}
	where $\mathrm{sg}(\cdot)$ denotes stop-gradient. The overall agent loss is
	\begin{equation}
		\mathcal{L}_{\mathrm{agent}}
		=
		\mathcal{L}_{\mathrm{policy}}
		+
		c_v\mathcal{L}_{\mathrm{value}}
		-
		c_e\mathcal{H}
		\left(
		\pi_{\theta}^{\mathrm{mask}}
		\right),
		\label{eq:agent_loss}
	\end{equation}
	where $\mathcal{H}(\cdot)$ is the entropy term over the valid action set, and $c_v$ and $c_e$ balance value regression and exploration.
	
	After the agent update, the executed feature-extraction modules, invoked transition interfaces, auxiliary head, FPN neck, and detection head are optimized by task-level losses:
	\begin{equation}
		\mathcal{L}_{\mathrm{task}}
		=
		\mathcal{L}_{\mathrm{det}}
		+
		\lambda_{\mathrm{aux}}\mathcal{L}_{\mathrm{aux}}.
		\label{eq:task_loss}
	\end{equation}
	Here, $\mathcal{L}_{\mathrm{det}}$ is the standard oriented detection loss, and $\mathcal{L}_{\mathrm{aux}}$ supervises the auxiliary objectness and feature-quality predictions. The sampled module indices are treated as discrete routing decisions during this task update. Therefore, task gradients update only the executed modules and detection-related heads, while PPO gradients update only the routing agent. This decoupled optimization avoids unstable gradient interference between discrete policy learning and continuous feature representation learning.

	\section{Experimental Details}
	\label{sec:experimental_details}
	
	\subsection{Datasets and Implementation Details}
	
	To evaluate the effectiveness of BMCR across diverse remote sensing scenarios, we conduct experiments on three widely used oriented object detection benchmarks: DOTA, DIOR-R, and FAIR1M. These datasets cover different imaging conditions, object scales, category distributions, and scene complexities, providing a challenging testbed for adaptive backbone composition.
	
	\begin{itemize}
		\item \textbf{DOTA-v1.0}~\cite{Xia_2018_CVPR} and \textbf{DOTA-v1.5}~\cite{2022dota} are used for large-scale oriented object detection. Following common practice, large images are cropped into $1024 \times 1024$ patches with a 200-pixel overlap. BMCR is evaluated under single-scale and multi-scale settings, with scales of 0.5, 1.0, and 1.5 for the latter.
		
		\item \textbf{DIOR-R}~\cite{dior} is used to evaluate oriented object detection under diverse scene contexts and large intra-class variations. Following common settings, we use an input size of $800 \times 800$ for both training and testing.
		
		\item \textbf{FAIR1M-v1.0}~\cite{sun2022fair1m} is used to evaluate fine-grained oriented object detection in large-scale aerial imagery. The dataset contains dense object instances, subtle category differences, and substantial scale variations, making it suitable for assessing the adaptability of dynamic backbone composition. BMCR follows the same multi-scale configuration as DOTA for training and testing.
	\end{itemize}
	
	We compare BMCR with representative oriented object detectors, including EMO2-DETR~\cite{EMO2-DETR}, Oriented R-CNN~\cite{Xie_2021_ICCV}, OrientedFormer~\cite{zhao2024orientedformer}, and other recent methods covering CNN-based, Transformer-based, and hybrid architectures. We follow the reported training and evaluation protocols whenever available. Unless otherwise specified, BMCR uses ImageNet-pretrained backbone modules~\cite{russakovsky2015imagenet}, whereas some baselines adopt remote-sensing pretraining or task-specific initialization~\cite{wang2022empirical}. Results are summarized in Table~\ref{tab:merged_sota}.
	
	All experiments are conducted on six NVIDIA RTX 6000 Ada GPUs. The maximum number of routing steps is set to $T_{\max}=12$, providing sufficient routing flexibility while keeping the inference cost comparable to common backbone-level baselines. BMCR is trained using the proposed two-stage AMCO protocol. In the random-path warm-up phase, the policy network is frozen and the model is trained for 50 epochs by uniformly sampling actions from the masked valid action set. This stage warms up the heterogeneous toolbox, invoked transition interfaces, detection heads, and the value network. The auxiliary feature-quality score $q_t$ is computed from normalized localization, classification, and objectness terms, with relative weights of 0.6, 0.4, and 0.2, respectively, followed by score normalization.
	
	In the PPO routing optimization phase, the model is further trained for 100 epochs. Trajectories are collected with the old policy and current toolbox parameters, after which the policy and value networks are updated using the agent loss, including the clipped PPO objective, value regression, and entropy regularization. The clipping threshold is set to $\epsilon=0.2$, the discount factor to $\gamma=0.99$, and the GAE coefficient to $\lambda_{\mathrm{GAE}}=0.95$. The value-loss and entropy weights are set to $c_v=0.5$ and $c_e=0.01$, respectively. The transition coefficient $\lambda_{\mathrm{tr}}(n)$ and objectness-prior coefficient $\mu(n)$ are linearly decayed during the first 20 epochs of the PPO phase, where $N_{\mathrm{tr}}=N_{\mathrm{prior}}=20$.
	
	Both stages use the AdamW optimizer with a weight decay of $5\times10^{-4}$. The initial learning rate is set to $1\times10^{-4}$ and decayed to $1\times10^{-5}$ at the 120th epoch. For the reward function, the coefficients are set to $\lambda_{\mathrm{acc}}=1.0$, $\lambda_{\mathrm{cost}}=0.3$, $\lambda_{\mathrm{sw}}=0.2$, and $\lambda_{\mathrm{end}}=5.0$. Unless otherwise stated, the same optimization hyperparameters are used across datasets, while dataset-specific data processing and training schedules follow the corresponding benchmark protocols.

	\begin{table*}[t]
		\centering
		\fontsize{9pt}{10pt}\selectfont
		\renewcommand{\arraystretch}{1.2}
		\caption{
			Comparison with representative state-of-the-art oriented object detection methods on DOTA-v1.0, DOTA-v1.5, DIOR-R, and FAIR1M-v1.0 under single-scale (SS) and multi-scale (MS) settings. Best results in each column are highlighted in \textbf{bold}. FPS$_{1024}$ denotes the inference speed measured with $1024\times1024$ inputs.
		}
		\label{tab:merged_sota}
		\begin{tabular}{l l c c c c c c c }
			\hline
			\multirow{3}{*}{\textbf{Method}} 
			& \multirow{3}{*}{\textbf{Backbone}} 
			& \multicolumn{2}{c}{\textbf{DOTA-v1.0}} 
			& \multicolumn{2}{c}{\textbf{DOTA-v1.5}} 
			& \textbf{DIOR-R} 
			& \textbf{FAIR1M-v1.0} 
			& \textbf{Speed} \\
			\cline{3-9}
			& & SS & MS & SS & MS & SS & MS & Size$_{1024}$ \\
			& & mAP↑ & mAP↑ & mAP↑ & mAP↑ & mAP↑ & mAP$_F$↑ & FPS↑ \\
			
			\hline
			\multirow{2}{*}{EMO2-DETR~\cite{EMO2-DETR}}  & ResNet-50  \cite{he2016resnet} & 70.91  & 76.23  & 61.67  & 71.79 & - & - & - \\
			& Swin-T ~\cite{liu2021swin} & 72.32  & 78.46  & -  & - & - & - & - \\
			\hline
			\multirow{10}{*}{Oriented R-CNN~\cite{Xie_2021_ICCV}} & ResNet-50 \cite{he2016resnet} & 73.50 & 80.87 & 66.86 & 72.20 & 67.87 & 41.60 & \textbf{68.5}  \\
			& ResNet-101 \cite{he2016resnet} & 75.82 & 80.52 & 67.50 & 73.67 & 67.92 & 42.32 & 52.4  \\
			& Swin-T~\cite{liu2021swin} & 75.95 & 80.37 & 66.45 & 75.24 & 68.03 & 45.54 & 46.5  \\
			& LSK-T~\cite{li2023lsknet} & 77.60  & 81.37 & 66.71 & 75.52 & 68.42 & 46.50  &  49.5  \\
			& LSK-S~\cite{li2023lsknet} & 78.02  & 81.64 & 70.26 & 75.58 & 66.88 &  47.85 &   44.8 \\
			& PKINet-S~\cite{cai2024pkinet} & 78.39  & 81.06 & 70.69 & 76.61  & 68.47 & 45.27 &  51.4 \\
			& ViTAEv2~\cite{wang2022empirical} & 77.72 & 81.24 & 70.17 & 75.52 & 67.37 & 45.80 &  24.3\\
			& ViTAE-B+RVSA~\cite{wang2022advancing}   & 78.99 & 81.18  & 70.39 & 75.37 & 71.06 & 46.13 & 20.5\\
			& ARC-R50~\cite{pu2023adaptive} & 77.35 & 80.93 & 71.42 & 77.33 & 69.70 & 45.94 & 55.2  \\
			& ARC-R101~\cite{pu2023adaptive} & 77.71 & 81.11 & 71.68 & 78.51 & 69.87 & 45.24 & 51.3  \\
			\hline
			\multirow{4}{*}{OrientedFormer~\cite{zhao2024orientedformer}} & ResNet-50  \cite{he2016resnet} & 75.37  & 79.06 & 67.06 & 73.41 & 67.28 & 40.17 & 61.3 \\
			& ResNet-101  \cite{he2016resnet} & 75.91  & 79.36 & 69.24 & 74.48 & 68.84 & 39.80 & 51.2 \\
			& Swin-T~\cite{liu2021swin} & 75.88  & 78.86 & 70.12 & 76.08 & 71.09 & 40.07 & 48.6 \\
			& LSK-T~\cite{li2023lsknet} & 77.13  & 79.61 & 71.57 & 77.61 & 65.05 & 42.56 &  47.8\\
			\hline
			\multirow{2}{*}{RoI Trans.~\cite{roi-transformer}}  & Swin-T~\cite{liu2021swin} & 74.61 & 79.05 & 70.27 & 75.57 & 66.78 & 35.29 & 56.4  \\
			& ViT-G~\cite{cha2024billion}  & 77.10 & 81.94 & - & - & 69.43 &- & 21.3  \\
			\hline
			RoI-Trans.-KFIoU \cite{yang2022kfiou} & Swin-T~\cite{liu2021swin} & 76.94 & 80.93 & 70.46 & 74.28 &68.92& 46.15& 50.0 \\
			R3Det-KFIoU \cite{yang2022kfiou} & ResNet-152 \cite{he2016resnet} & 77.07 & 81.03 & 71.89 & 75.25 &69.75& 46.47&41.6  \\
			\hline
			RQFormer~\cite{RQFormer} & ResNet-50 \cite{he2016resnet} & 75.04  & 80.92  & 67.43 & 71.58  & 67.31 & 41.28 & 66.3 \\
			\hline
			\multirow{3}{*}{UniconDet~\cite{10879059}} & ResNet-50 \cite{he2016resnet} & 78.40 & - & - & - & - & - & - \\
			& ResNet-101 \cite{he2016resnet} & 78.45 & - & - & - & - & - & - \\
			& ResNet-152 \cite{he2016resnet} & 78.67 & - & - & - & - & - & - \\
			\hline
			\multirow{2}{*}{ARS-DETR~\cite{zeng2024ars}} & ResNet-50 \cite{he2016resnet} & 74.16 & 78.17 & 68.81 & 74.29 & 66.12 & 45.08 & 61.4  \\
			& Swin-T \cite{liu2021swin} & 75.47 & 78.09 & 68.29 & 73.97 & 68.13 & 46.13 & 55.1 \\
			\hline
			Strip R-CNN~\cite{yuan2025strip}  & StripNet-S~\cite{yuan2025strip} & 78.37  & 81.58 & 72.27 & 76.15  & 70.07 & 48.26 &66.3\\
			\rowcolor{gray!20} 
			Oriented R-CNN~\cite{Xie_2021_ICCV} & {BMCR (Ours)}  & \textbf{79.31}  & \textbf{82.13} & \textbf{73.41}  & \textbf{79.40}  & \textbf{71.86}  & \textbf{49.15} & 63.9\\
			\hline
		\end{tabular}
	\end{table*}

	\begin{figure*}[t]
		\centering
		\includegraphics[width=0.99\textwidth]{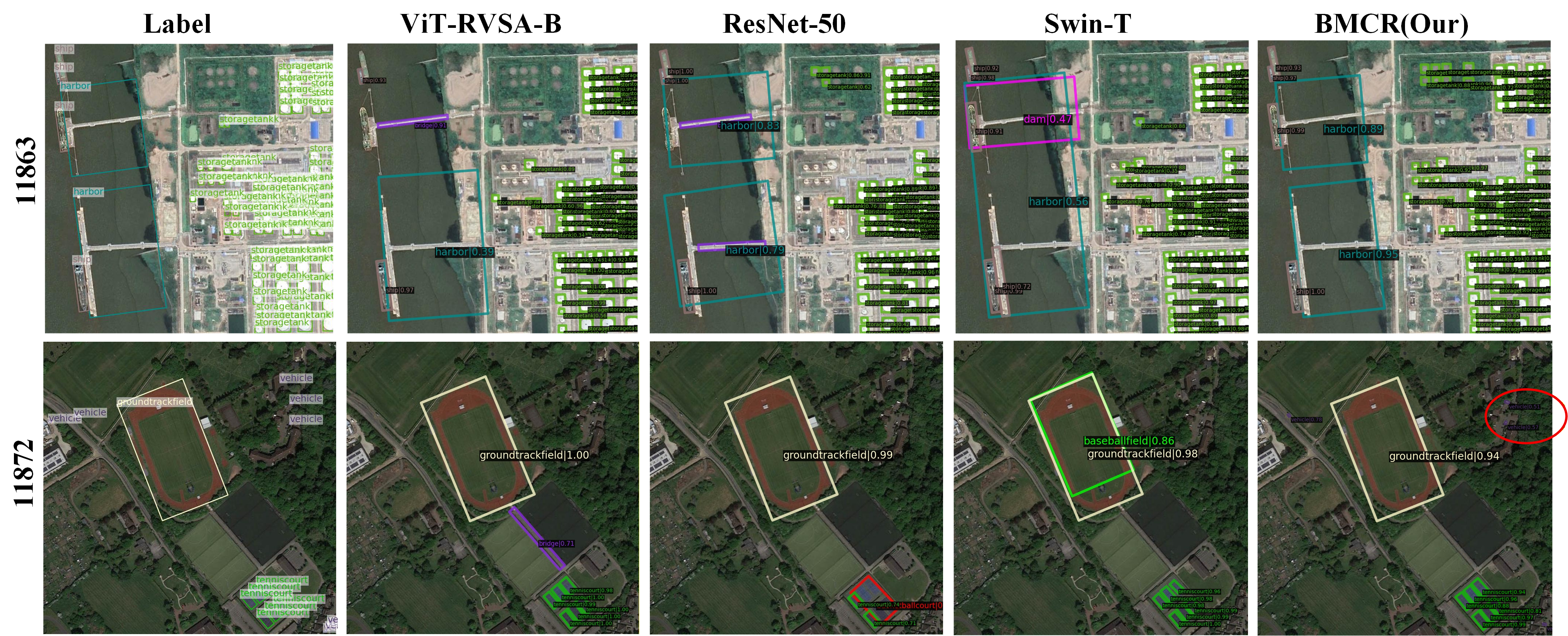}
		\caption{
			Qualitative comparison of oriented object detection results on large remote sensing images.
			Compared with representative static backbones, BMCR yields more accurate detections under dense object distributions, sparse layouts, and complex background clutter.
		}
		\label{fig:qualitative_detection}
	\end{figure*}
	
	\paragraph{Overall Comparison}
	Table~\ref{tab:merged_sota} compares BMCR with representative oriented object detection methods covering CNN-based, Transformer-based, hybrid, and DETR-style architectures. Overall, BMCR achieves the best accuracy on all reported benchmark columns, including DOTA-v1.0, DOTA-v1.5, DIOR-R, and FAIR1M-v1.0, while maintaining competitive inference speed. Compared with static backbone variants under Oriented R-CNN, BMCR consistently improves accuracy across datasets, indicating that adaptive backbone composition provides benefits beyond simply increasing backbone capacity. Compared with specialized detectors and heavy Transformer backbones, BMCR also achieves a more favorable accuracy--efficiency balance, suggesting that input-adaptive routing can allocate computation more effectively across heterogeneous remote sensing scenes with diverse object scales and spatial layouts.
	
	\paragraph{Results on DOTA}
	Table~\ref{tab:merged_sota} reports the comparison results on DOTA-v1.0 and DOTA-v1.5 under single-scale (SS) and multi-scale (MS) testing settings. DOTA is particularly challenging due to its large image size, dense object layouts, arbitrary object orientations, and substantial scale variations. These characteristics require detectors to capture fine-grained local structures for dense small objects while maintaining sufficient context for complex aerial scenes.
	
	BMCR achieves the best performance on both DOTA versions. On DOTA-v1.0, BMCR obtains 79.31\% and 82.13\% mAP$_{50}$ under SS and MS settings, respectively, outperforming strong Oriented R-CNN variants with advanced backbones such as LSK-S and ViTAE-B+RVSA. On DOTA-v1.5, which contains more challenging small-object instances, BMCR achieves 73.41\% and 79.40\% mAP$_{50}$ under SS and MS settings, improving over the best competing results by 1.14 and 0.89 percentage points. These gains show that adaptive module composition is effective for dense small objects and large scale variations, enabling BMCR to construct inference paths that better fit complex aerial scenes.
	
	\paragraph{Results on DIOR-R}
	As shown in Table~\ref{tab:merged_sota}, BMCR achieves 71.86\% mAP$_{50}$ under the single-scale setting, obtaining the best result among all compared methods. DIOR-R contains diverse scene contexts and large intra-class variations, making it necessary to balance local detail extraction and semantic discrimination. BMCR outperforms representative CNN- and Transformer-based baselines, including OrientedFormer with Swin-T and RoI Trans. with ViT-G, indicating that adaptive composition of reusable modules provides stronger feature flexibility than a single static backbone. The qualitative results in Fig.~\ref{fig:qualitative_detection} further support this observation. In the first image, 11863.jpg, several compared methods produce false alarms on confusing dam- and bridge-like structures, whereas BMCR suppresses these erroneous detections. In the second image, 11872.jpg, BMCR detects more small vehicles in dense and cluttered layouts, showing its advantage in preserving fine-grained object cues. These results demonstrate that BMCR generalizes well to diverse oriented detection scenarios beyond DOTA-style benchmarks.

	\paragraph{Results on FAIR1M-v1.0}
	FAIR1M-v1.0 is a large-scale fine-grained oriented object detection benchmark with dense annotations and subtle category differences. Under the multi-scale testing setting, BMCR achieves 49.15\% mAP$_F$, outperforming the strongest compared baseline by 0.89 percentage points. The improvement over strong backbones such as StripNet-S and ViTAE-B+RVSA indicates that BMCR is effective for fine-grained category discrimination under dense instance distributions. This advantage is consistent with the design motivation of dynamic module composition: different scenes and categories may require different combinations of local detail modeling, long-range context aggregation, and feature recalibration.
	
	\paragraph{Inference Efficiency}
	In addition to detection accuracy, we report inference speed to evaluate practical efficiency. FPS is measured on a single NVIDIA RTX 6000 Ada GPU using $1024 \times 1024$ input images, as shown in Table~\ref{tab:merged_sota}. BMCR achieves 63.9 FPS, which is slightly lower than the fastest lightweight baselines, such as Oriented R-CNN with ResNet-50 and RQFormer with ResNet-50, but remains competitive among high-accuracy methods. Compared with the heavy ViT-G based RoI Trans. baseline, BMCR improves the speed from 21.3 FPS to 63.9 FPS while achieving higher accuracy on the commonly reported benchmarks. These results indicate that BMCR provides a favorable accuracy--efficiency trade-off: although dynamic routing introduces additional decision overhead, the overall inference speed remains competitive due to adaptive module composition.

	\subsection{Comparative Study}
	
	We further conduct a controlled study on DIOR-R to evaluate the proposed RL-driven module combination strategy. Unlike the main comparison in Table~\ref{tab:merged_sota}, all methods use the same Oriented R-CNN head and ImageNet-pretrained backbone initialization~\cite{russakovsky2015imagenet}. This setting reduces the influence of external factors and enables a direct comparison of backbone composition mechanisms.
	
	We compare BMCR with both static backbone baselines and alternative decision strategies. The static baselines include representative CNN, Transformer, and hybrid backbones under the same detection head. To evaluate the role of RL-based routing, we further introduce three decision-strategy baselines using the same module toolbox and interface design as BMCR:
	
	\begin{itemize}
		\item \textbf{Step-wise Greedy}: This baseline selects the locally best module at each routing step according to the auxiliary feature-quality score. Unlike BMCR, it does not perform long-horizon policy learning and therefore ignores the delayed effect of current module choices on subsequent routing decisions.
		
		\item \textbf{Random Strategy}: This baseline samples 10,000 structurally valid 12-step paths from the module toolbox. The best path is selected on the validation set and then evaluated on the test set. This setting evaluates whether simple random exploration can approximate effective module composition in the large discrete search space.
		
		\item \textbf{Dynamic Routing}: This baseline replaces the PPO-based policy with a differentiable soft-routing strategy based on Top-$K$ Gumbel-Softmax~\cite{kool2019stochastic}. During training, the top-$K$ high-weight modules are aggregated at each step with $K=8$, while at inference the module with the maximum routing probability is selected at each of the 12 steps.
	\end{itemize}
	
	As reported in Table~\ref{tab:compare}, fixed backbones provide stable but limited performance under the controlled Oriented R-CNN setting. Stronger Transformer-based and hybrid backbones, including Swin-B, SwinV2-B, ViTAEv2-B, and ViT-RVSA-B, improve over ResNet variants, showing the benefit of stronger feature representations for DIOR-R. However, these static architectures apply the same computation path to all images, which limits their adaptability to heterogeneous remote sensing scenes with diverse object layouts.
	
	The adaptive decision strategies exhibit different behaviors. Step-wise Greedy achieves 68.82\% mAP$_{50}$, but its locally optimal decisions ignore the delayed influence of early module choices. Random Strategy obtains 60.90\% mAP$_{50}$, suggesting that random search is inefficient in the large discrete routing space. Dynamic Routing reaches 68.95\% mAP$_{50}$, indicating that differentiable soft routing can learn useful module preferences. However, its mismatch between soft multi-module aggregation during training and hard single-path selection during inference limits the final detection performance.
	
	BMCR achieves the best accuracy, reaching 71.86\% mAP$_{50}$ while maintaining 70.4 FPS. Although ResNet-50 remains the fastest static baseline, BMCR achieves the highest inference speed among adaptive decision strategies and substantially improves their detection accuracy. These results confirm that PPO-based long-horizon routing is more effective than greedy selection, random search, and soft differentiable routing for learning sample-specific backbone composition.
	
	\begin{table}[t]
		\centering
		\caption{Controlled comparative study on the DIOR-R dataset. All methods use the same Oriented R-CNN detection head and ImageNet-pretrained initialization. FPS is measured with $800 \times 800$ input images.}
		\fontsize{9pt}{10pt}\selectfont
		\begin{tabular}{lcc}
			\hline
			\textbf{Method / Backbone} & \textbf{FPS}↑ & \textbf{mAP$_{50}$ (\%)}$\uparrow$ \\
			\hline
			ResNet-50              & \textbf{78.9} & 67.87 \\
			ResNet-101             & 71.4          & 67.92 \\ 
			Swin-T                 & 66.4          & 68.03 \\
			Swin-S                 &  60.7         & 68.61 \\
			Swin-B                 & 55.2          & 69.21 \\
			SwinV2-T               & 56.2          & 67.93 \\
			SwinV2-S               & 59.8          & 68.21 \\
			SwinV2-B               & 55.6          & 69.33 \\
			ViT-B                  & 28.6          & 67.34 \\
			ViTAEv2-B              & 25.7          & 69.37 \\
			ViT-RVSA-B             & 24.6          & 71.06 \\
			\hline
			DynamicViT   & 56.9          & 63.01 \\  
			Step-wise Greedy       &  51.4         & 68.82 \\ 
			Random Strategy        & 53.9          & 60.90 \\ 
			Dynamic Routing        & 55.1          & 68.95 \\
			\rowcolor{gray!20} 
			BMCR (Ours)            & 70.4          & \textbf{71.86} \\ 
			\hline
		\end{tabular}
		\label{tab:compare}
	\end{table}

	\subsection{Ablation Study}
	
	\begin{table}[t]
		\centering
		\caption{Ablation study of BMCR's key components on the DIOR-R dataset. $\Delta$ denotes the mAP$_{50}$ difference relative to the full BMCR model.}
		\fontsize{9pt}{10pt}\selectfont
		\begin{tabular}{lcc}
			\hline
			\textbf{Configuration} & \textbf{mAP$_{50}$ (\%)}$\uparrow$ & \textbf{$\Delta$} \\
			\hline
			CNN-only Toolbox & 68.20 & -3.66 \\
			ViT-only Toolbox & 64.27 & -7.59 \\
			\hline
			BMCR (Full Model) & \textbf{71.86} & 0.00 \\
			w/o OT-based Interface & 67.15 & -4.71 \\
			w/o MSP Encoder & 69.31 & -2.55 \\
			w/o Random-Path Warm-up & 65.25 & -6.61 \\
			w/o Auxiliary Spatial Prior & 70.58 & -1.28 \\
			w/o Auxiliary Detection Head & 69.83 & -2.03 \\
			\hline
		\end{tabular}
		\label{tab:ablation}
	\end{table}
	
	Table~\ref{tab:ablation} reports the ablation results on DIOR-R. The full BMCR model achieves 71.86\% mAP$_{50}$, and each variant is described below.
	
	We first examine the composition of the module toolbox. In the CNN-only variant, the action space is restricted to modules derived from ResNet backbones and the termination action, while the ViT-only variant retains only modules from ViT, Swin, SwinV2, ViTAE, and ViTAE-RVSA. Both keep the same total number of module entries and maximum routing steps, and follow the same AMCO training protocol. The CNN-only and ViT-only toolboxes obtain 68.20\% and 64.27\% mAP$_{50}$, dropping by 3.66 and 7.59 points, respectively, confirming that heterogeneous modules provide complementary local and global representations crucial for diverse remote sensing scenes.
	
	The remaining ablations evaluate the transition interface, state encoder, and training strategy. When the OT-based interface is removed (w/o OT-based Interface), all cross-family transitions are handled by the reduction interface using bilinear interpolation and channel resampling. This causes a 4.71-point drop to 67.15\%, indicating that simple spatial resampling is insufficient to align CNN grids and ViT tokens. Replacing the multi-scale pyramid state encoder with a single-scale feature (w/o MSP Encoder) reduces mAP$_{50}$ to 69.31\%, showing the benefit of multi-scale observations for handling large scale variations. In terms of training, removing the random-path warm-up phase and starting directly with PPO optimization (w/o Random-Path Warm-up) leads to a sharp decline to 65.25\% mAP$_{50}$, a 6.61-point loss that underscores the importance of the warm-up stage in stabilizing early joint optimization. Excluding the auxiliary spatial prior from the agent state (w/o Auxiliary Spatial Prior) yields 70.58\%, demonstrating that the predicted objectness map provides useful localization guidance. Removing the auxiliary detection head entirely (w/o Auxiliary Detection Head) further reduces performance to 69.83\%, since the value network then relies solely on sparse RL returns without dense auxiliary supervision. Collectively, these results validate the contributions of heterogeneous module composition, the OT-based interface, multi-scale state encoding, and the AMCO training components.

	\subsection{Trade-off between Efficiency and Accuracy}
	
	To evaluate the flexibility of BMCR in balancing detection accuracy and inference efficiency, we conduct a trade-off analysis by varying the computational penalty coefficient $\lambda_{\mathrm{cost}}$ in Eq.~\eqref{eq:reward_function}. A larger $\lambda_{\mathrm{cost}}$ encourages the routing agent to select shorter or less expensive module paths, whereas a smaller value allows the agent to allocate more computation for higher detection accuracy. In this experiment, all training schedules and evaluation protocols are kept unchanged, and only $\lambda_{\mathrm{cost}}$ is adjusted to analyze its impact on inference latency and mAP$_{50}$.
	
\begin{figure}[t]
		\centering
		\includegraphics[width=1\columnwidth]{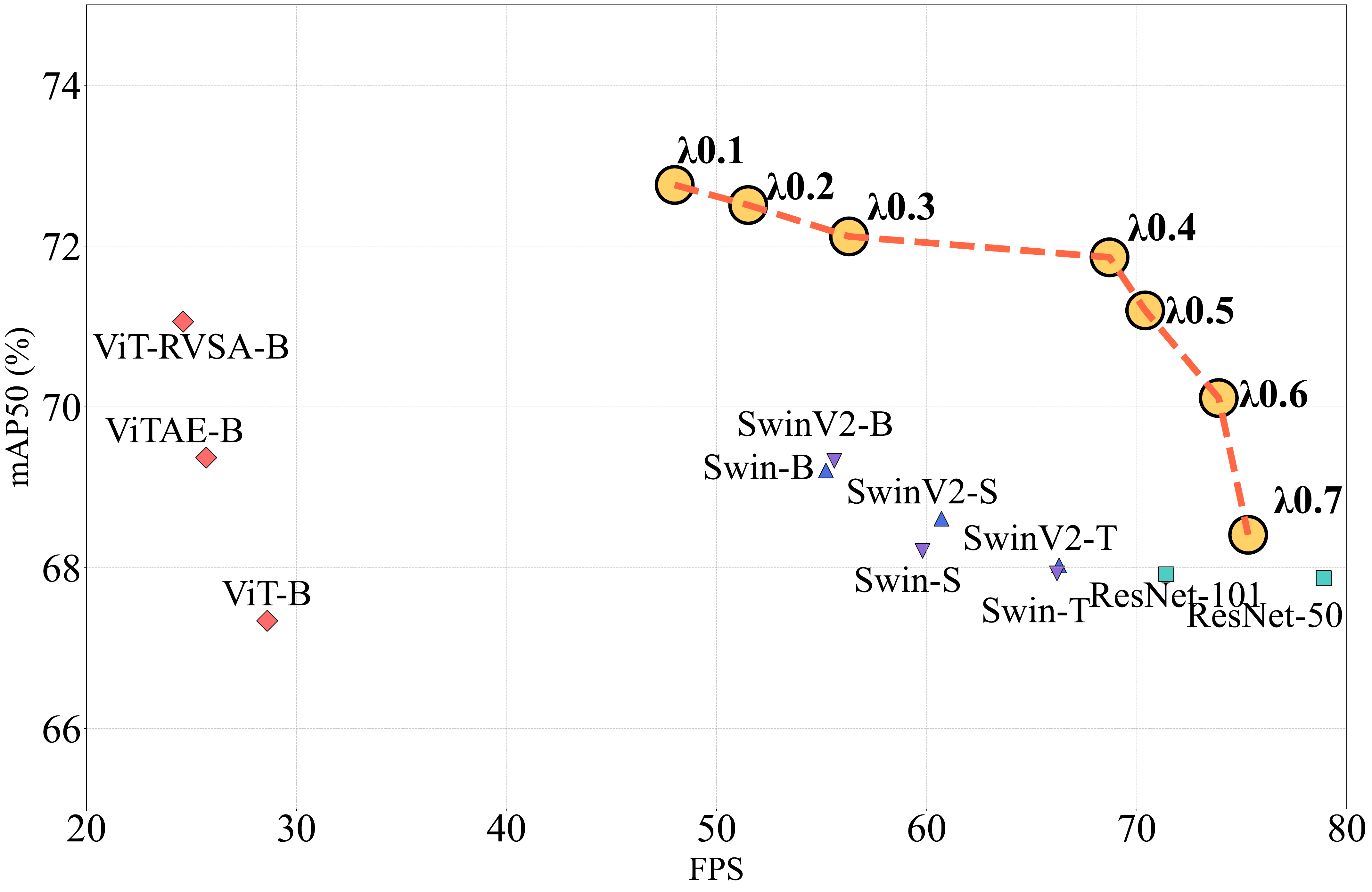}
		\caption{Accuracy--efficiency trade-off of BMCR under different computational penalty coefficients $\lambda_{\mathrm{cost}}$. By increasing $\lambda_{\mathrm{cost}}$, the routing agent is encouraged to select more efficient inference paths, leading to lower latency with moderate accuracy degradation.}
		\label{fig:balance}
	\end{figure}

	As shown in Fig.~\ref{fig:balance}, BMCR provides a smooth accuracy--efficiency trade-off under different computational penalty settings. When $\lambda_{\mathrm{cost}}$ is small, the agent tends to select more expressive routing paths, achieving the highest mAP$_{50}$ of approximately 73.5\%, but at the cost of higher inference latency. As $\lambda_{\mathrm{cost}}$ increases, the agent progressively favors more efficient paths, reducing the inference time to 42.3 ms while maintaining a competitive mAP$_{50}$ of 71.4\%. This result indicates that BMCR can adapt its average active computation according to deployment requirements, providing flexible operating points for scenarios with different accuracy and latency constraints.
	
	\begin{figure*}[!t]
		\centering
		\includegraphics[width=0.98\textwidth]{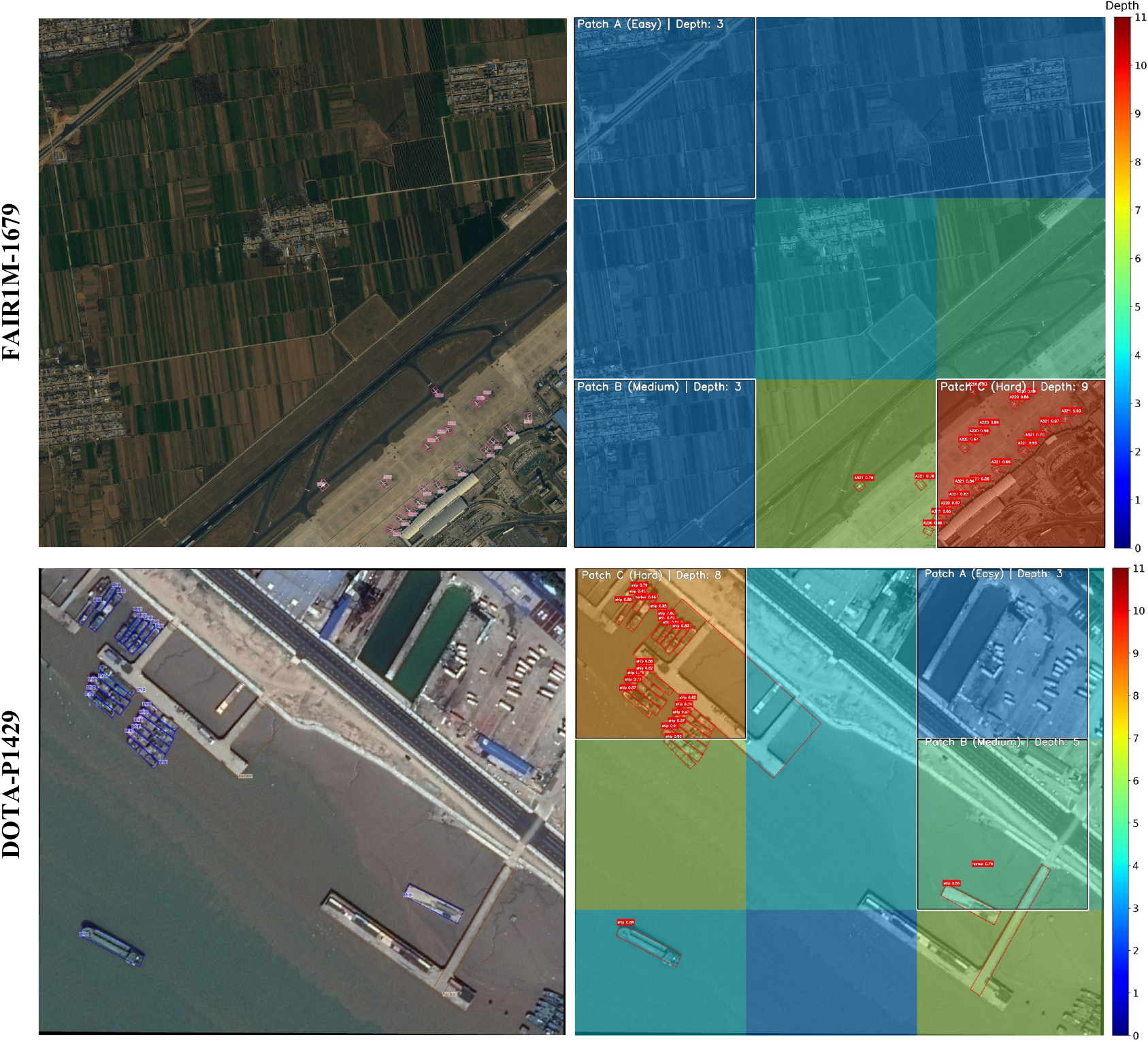}
		\caption{
			Visualization of BMCR's dynamic routing behavior on large-scale remote sensing imagery. Cold colors indicate shallow routing in simple regions, while warm colors denote deeper computation in complex or target-dense areas.
		}
		\label{fig:routing_vis}
	\end{figure*}
	\section{Analysis and Discussion}
	
	\subsection{Influence of Maximum Decision Steps}
	
\begin{figure}[t]
		\centering
		\includegraphics[width=\columnwidth]{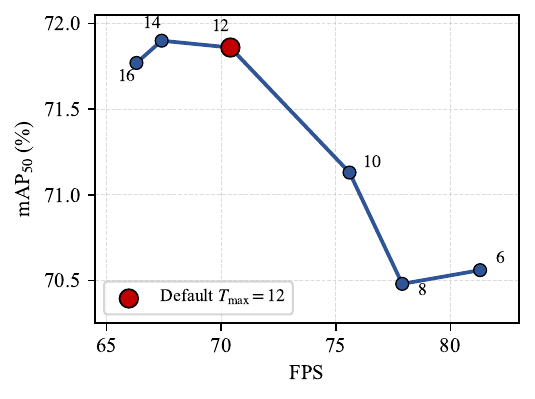}
		\caption{
			Accuracy--efficiency trade-off under different maximum routing horizons $T_{\max}$.
			Each point denotes one setting on the DIOR-R validation set.
			$T_{\max}=12$ is used as the default setting due to its better balance between accuracy and speed.
		}
		\label{fig:tmax_tradeoff}
	\end{figure}

	We study the effect of the maximum routing horizon $T_{\max}$ on the accuracy--efficiency trade-off of BMCR. All other settings are kept unchanged. Fig.~\ref{fig:tmax_tradeoff} reports the mAP$_{50}$ and FPS on the DIOR-R validation set.
	
	Increasing $T_{\max}$ from 6 to 12 improves mAP$_{50}$ from 70.56\% to 71.86\%, showing that a longer horizon allows the agent to compose more informative paths. However, larger horizons bring little extra gain but reduce inference speed. When $T_{\max}$ increases from 12 to 14, mAP$_{50}$ improves by only 0.04 points, while FPS drops from 70.4 to 67.4. Further increasing $T_{\max}$ to 16 slightly decreases accuracy, suggesting that overly long paths may introduce redundant transformations. Therefore, we use $T_{\max}=12$ as the default setting for a better accuracy--efficiency balance.
	
	\subsection{Analysis on Large-Scale Imagery}

	\begin{figure}[t]
		\centering
		\includegraphics[width=\columnwidth]{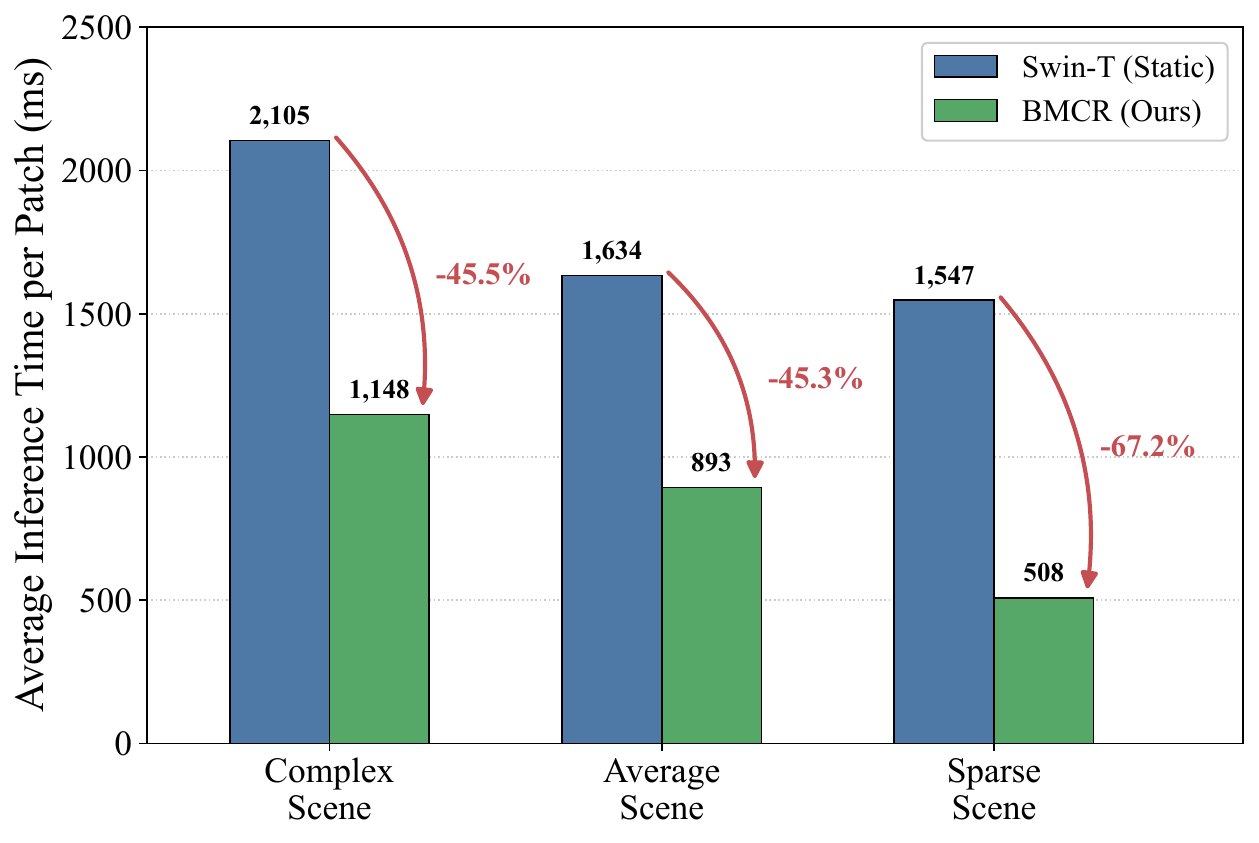}
		\caption{
			Average inference time for large-image inference.
			BMCR reduces the inference time across complex, average, and sparse scenes, with the largest gain in sparse scenes.
		}
		\label{fig:scene_latency}
	\end{figure}
	
	A key advantage of BMCR is its ability to allocate computation adaptively during large-scale remote sensing inference. Very-high-resolution aerial and satellite images are commonly processed by sliding windows, where each image is decomposed into many local patches. However, the semantic complexity of these patches is highly uneven. Many patches contain homogeneous backgrounds such as water, farmland, bare land, or runways, while only a subset contains dense targets, complex structures, or high-frequency details. Static backbones apply the same full computation to all patches, leading to considerable redundancy in simple regions.
	
	BMCR addresses this issue through patch-level dynamic routing. Instead
	of assigning an identical computation budget to every patch, BMCR
	determines the routing depth according to local visual complexity. As
	shown in Fig.~\ref{fig:routing_vis}, easy patches dominated by
	homogeneous backgrounds are assigned shallow routes, such as Patch~A
	with a routing depth of 3. Medium-complexity regions, such as Patch~B,
	receive moderate computation, while hard regions containing dense
	targets or complex structures, such as Patch~C, activate deeper routes
	with depths of 8 or 9. Rather than reducing computation uniformly,
	BMCR reallocates it toward regions requiring stronger representations.

	The quantitative comparison in Fig.~\ref{fig:scene_latency} further verifies this advantage. Compared with the static Swin backbone, BMCR reduces the average inference time from 2105 ms to 1148 ms in complex scenes, from 1634 ms to 893 ms in average scenes, and from 1547 ms to 508 ms in sparse scenes. These correspond to relative reductions of 45.5\%, 45.3\%, and 67.2\%, respectively. The largest gain appears in sparse scenes, where many patches can be processed with shorter paths because they contain limited object-related information.
	
	These results demonstrate that BMCR is particularly suitable for large-image remote sensing applications. By allocating deeper computation to complex regions and reducing unnecessary processing in homogeneous areas, BMCR better matches the intrinsic spatial heterogeneity of remote sensing imagery. This property enables efficient large-scale inference while preserving the representation capacity required for dense targets and complex geographic structures.
	
	\subsection{Component-Level Inference Latency}
	\begin{figure}[!t]
		\centering
		\includegraphics[width=0.5\textwidth]{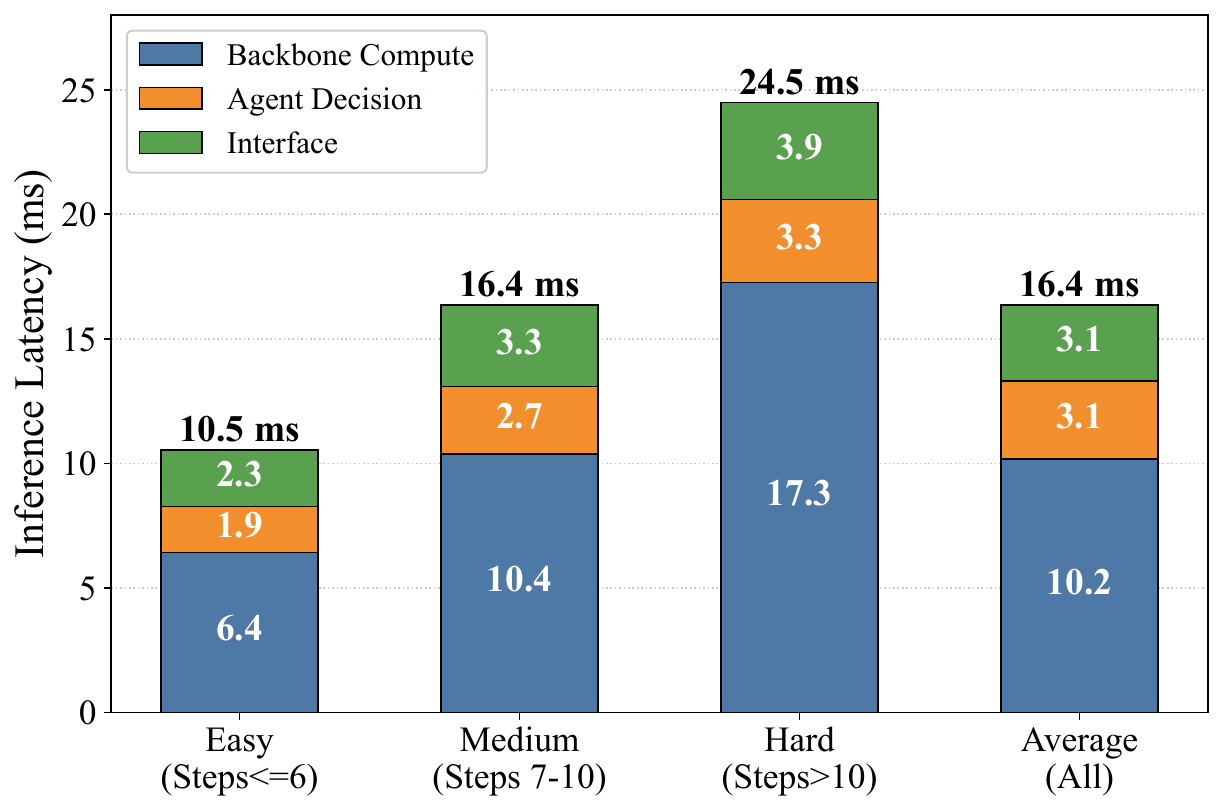}
		\caption{Inference latency breakdown of BMCR under different routing complexities. 
			Latency is decomposed into backbone computation, agent decision, and interface 
			overhead. Backbone computation remains the dominant cost, while the routing 
			agent and interface introduce limited additional latency, yielding an average 
			inference latency of 16.4 ms.}
		\label{fig:latency_breakdown}
	\end{figure}
	
	To further analyze the runtime cost of BMCR, we decompose inference latency into backbone computation, agent decision, and interface transformation. As shown in Fig.~\ref{fig:latency_breakdown}, the overall latency increases from 10.5 ms for easy samples to 24.5 ms for hard samples, since harder samples require more routing steps and execute more backbone modules. Backbone computation remains the dominant cost, accounting for 6.4 ms, 10.4 ms, and 17.3 ms on easy, medium, and hard samples, respectively. By contrast, the agent decision cost is limited to 1.9--3.3 ms, and the interface transformation cost is limited to 2.3--3.9 ms.
	
	On average, BMCR requires 16.4 ms per patch, with 10.2 ms spent on backbone computation and only 3.1 ms each on agent decision and interface transformation. This indicates that the dynamic routing mechanism and heterogeneous transition interface introduce moderate overhead, while most latency is still dominated by actual feature extraction. Therefore, BMCR can achieve adaptive computation allocation without making routing or feature alignment the runtime bottleneck.
	
	\subsection{Training Overhead and Memory Analysis}
	
	Table~\ref{tab:training_cost} reports the training wall-clock time and
	peak GPU memory of BMCR and two representative static backbones on
	DIOR-R. All measurements use NVIDIA RTX 6000 Ada GPUs with a total
	batch size of 12 and a 150-epoch schedule. BMCR's full AMCO protocol
	requires approximately 48.5 hours and 46.2 GB peak memory, compared
	with 6.2 h / 16.1 GB for ResNet-50 and 8.8 h / 17.7 GB for Swin-T.
	
	The overhead arises from two sources: (i) the random-path warm-up
	(~13.2 h) executes diverse valid paths to initialize the heterogeneous
	toolbox under a frozen policy; (ii) the PPO stage (~35.3 h) collects
	on-policy trajectories and stores intermediate features for
	policy-gradient updates, a cost inherent to RL-based optimization.
	Notably, the trained toolbox and transition interfaces are
	transferable across datasets. When fine-tuning a DOTA-pretrained BMCR
	on DIOR-R, the warm-up stage is skipped entirely because the modules
	are already well-initialized, reducing training time from 48.5 h to
	12.1 h (75\% reduction). This demonstrates that the initial cost is a
	one-time investment amortizable across downstream tasks.
	
	The elevated peak memory stems from jointly maintaining multiple
	backbone families in the toolbox during training. This cost is
	confined to the training phase: at inference, only the modules along
	the selected path are activated, so per-sample memory is comparable to
	a single static backbone. The overall training cost is comparable to
	other RL-based dynamic networks requiring on-policy rollouts and
	substantially lower than large-scale NAS methods. Further reduction
	via off-policy algorithms and efficient toolbox sharing is left as
	future work.

	\begin{table}[t]
		\centering
		\caption{Training cost comparison on DIOR-R. All methods use the Oriented R-CNN head.}
		\fontsize{8.5pt}{9.5pt}\selectfont
		\setlength{\tabcolsep}{4pt}
		\renewcommand{\arraystretch}{1.08}
		\begin{tabular}{@{}lcc@{}}
			\hline
			\textbf{Method} & \textbf{Time (h)} & \textbf{Mem. (GB)} \\
			\hline
			ResNet-50 & 6.2  & 16.1 \\
			Swin-T    & 8.8  & 17.7 \\
			BMCR (AMCO) & 48.5 & 46.2 \\
			\quad Stage~1 & 13.2 & -- \\
			\quad Stage~2 & 35.3 & -- \\
			BMCR (Fine-tune) & 12.1 & 45.8 \\
			\hline
		\end{tabular}
		\label{tab:training_cost}
	\end{table}

	\section{Conclusion and Future Work}
	This paper presented BMCR, an RL-driven adaptive backbone composition framework for remote sensing oriented object detection. BMCR decomposes representative CNN and ViT backbones into reusable modules and assembles sample-specific inference paths through a module combination agent, while an OT-based interface supports CNN--ViT feature transitions and AMCO stabilizes the joint optimization of routing policies, heterogeneous modules, and detection heads. Extensive experiments on DOTA, DIOR-R, and FAIR1M demonstrate that BMCR improves detection performance over representative fixed-backbone and dynamic-routing baselines while maintaining a favorable accuracy--efficiency trade-off. Ablation studies further verify the effectiveness of the heterogeneous toolbox, OT-based interface, multi-scale state observation, auxiliary detection head, and random-path warm-up. Despite these advantages, BMCR still relies on a selected toolbox and involves more complex joint optimization than static backbones. Future work will explore more efficient toolbox expansion, incremental module registration, and runtime-controllable routing for deployment scenarios with varying latency, memory, and energy constraints.

	\small
	\bibliographystyle{IEEEtran}
	\bibliography{ref.bib}

@String(AAAI = {AAAI})

@InProceedings{Guo_2024_CVPR,
	author    = {Guo, Xin and Lao, Jiangwei and Dang, Bo and Zhang, Yingying and Yu, Lei and Ru, Lixiang and Zhong, Liheng and Huang, Ziyuan and Wu, Kang and Hu, Dingxiang and He, Huimei and Wang, Jian and Chen, Jingdong and Yang, Ming and Zhang, Yongjun and Li, Yansheng},
	title     = {SkySense: A Multi-Modal Remote Sensing Foundation Model Towards Universal Interpretation for Earth Observation Imagery},
	booktitle = {Proceedings of the IEEE/CVF Conference on Computer Vision and Pattern Recognition},
	month     = {June},
	year      = {2024},
	pages     = {27672--27683}
}

@ARTICLE{2023RingMo,
	author={Sun, Xian and Wang, Peijin and Lu, Wanxuan and Zhu, Zicong and Lu, Xiaonan and He, Qibin and Li, Junxi and Rong, Xuee and Yang, Zhujun and Chang, Hao and He, Qinglin and Yang, Guang and Wang, Ruiping and Lu, Jiwen and Fu, Kun},
	journal={IEEE Transactions on Geoscience and Remote Sensing},
	title={RingMo: A Remote Sensing Foundation Model With Masked Image Modeling},
	year={2023},
	volume={61},
	pages={1--22},
	doi={10.1109/TGRS.2022.3194732}}

@article{LI2022102926,
	title = {Deep learning in multimodal remote sensing data fusion: A comprehensive review},
	journal = {International Journal of Applied Earth Observation and Geoinformation},
	volume = {112},
	pages = {102926},
	year = {2022},
	issn = {1569-8432},
	doi = {10.1016/j.jag.2022.102926},
	author = {Jiaxin Li and Danfeng Hong and Lianru Gao and Jing Yao and Ke Zheng and Bing Zhang and Jocelyn Chanussot},
}

@inproceedings{hatamizadeh2023fastervit,
	author = {Hatamizadeh, Ali and Heinrich, Greg and Yin, Hongxu and Tao, Andrew and Alvarez, Jose M. and Kautz, Jan and Molchanov, Pavlo},
	booktitle = {International Conference on Learning Representations},
	editor = {B. Kim and Y. Yue and S. Chaudhuri and K. Fragkiadaki and M. Khan and Y. Sun},
	pages = {29368--29391},
	title = {FasterViT: Fast Vision Transformers with Hierarchical Attention},
	volume = {2024},
	year = {2024}
}

@inproceedings{he2016resnet,
	title={Deep residual learning for image recognition},
	author={He, Kaiming and Zhang, Xiangyu and Ren, Shaoqing and Sun, Jian},
	booktitle={Proceedings of the IEEE/CVF Conference on Computer Vision and Pattern Recognition},
	pages={770--778},
	year={2016},
	doi={10.1109/CVPR.2016.90}
}

@inproceedings{dosovitskiy2021an,
	title={An Image is Worth {16x16} Words: Transformers for Image Recognition at Scale},
	author={Dosovitskiy, Alexey and Beyer, Lucas and Kolesnikov, Alexander and Weissenborn, Dirk and Zhai, Xiaohua and Unterthiner, Thomas and Dehghani, Mostafa and Minderer, Matthias and Heigold, Georg and Gelly, Sylvain and Uszkoreit, Jakob and Houlsby, Neil},
	booktitle={International Conference on Learning Representations},
	year={2021},
}

@article{wei2024pathnet,
	title={PathNet: Path-selective point cloud denoising},
	author={Wei, Zeyong and Chen, Honghua and Nan, Liangliang and Wang, Jun and Qin, Jing and Wei, Mingqiang},
	journal={IEEE Transactions on Pattern Analysis and Machine Intelligence},
	volume={46},
	number={6},
	pages={4426--4442},
	year={2024},
	publisher={IEEE}
}

@article{yu2021path,
	title={Path-Restore: Learning network path selection for image restoration},
	author={Yu, Ke and Wang, Xintao and Dong, Chao and Tang, Xiaoou and Loy, Chen Change},
	journal={IEEE Transactions on Pattern Analysis and Machine Intelligence},
	volume={44},
	number={10},
	pages={7078--7092},
	year={2022},
}

@article{zhao2021battle,
	title={A Battle of Network Structures: An Empirical Study of {CNN}, Transformer, and {MLP}},
	author={Zhao, Yucheng and Wang, Guangting and Tang, Chuanxin and Luo, Chong and Zeng, Wenjun and Zha, Zheng-Jun},
	journal={arXiv preprint arXiv:2108.13002},
	year={2021}
}

@article{chen2025dynamicvis,
	title={DynamicVis: An efficient and general visual foundation model for remote sensing image understanding},
	author={Chen, Keyan and Liu, Chenyang and Chen, Bowen and Li, Wenyuan and Zou, Zhengxia and Shi, Zhenwei},
	journal={arXiv preprint arXiv:2503.16426},
	year={2025}
}

@inproceedings{cornebise2022open,
	title={Open high-resolution satellite imagery: The worldstrat dataset--with application to super-resolution},
	author={Cornebise, Julien and Or{\v{s}}oli{\'c}, Ivan and Kalaitzis, Freddie},
	booktitle={Advances in Neural Information Processing Systems},
	volume={35},
	pages={25979--25991},
	year={2022}
}

@article{schulman2017proximal,
	title={Proximal Policy Optimization Algorithms},
	author={Schulman, John and Wolski, Filip and Dhariwal, Prafulla and Radford, Alec and Klimov, Oleg},
	journal={arXiv preprint arXiv:1707.06347},
	year={2017}
}

@InProceedings{Xia_2018_CVPR,
	author={Xia, Gui-Song and Bai, Xiang and Ding, Jian and Zhu, Zhen and Belongie, Serge and Luo, Jiebo and Datcu, Mihai and Pelillo, Marcello and Zhang, Liangpei},
	booktitle={Proceedings of the IEEE/CVF Conference on Computer Vision and Pattern Recognition},
	title={{DOTA}: A Large-Scale Dataset for Object Detection in Aerial Images},
	year={2018},
	pages={3974--3983},
	doi={10.1109/CVPR.2018.00418}
}

@ARTICLE{2022dota,
	author={Ding, Jian and Xue, Nan and Xia, Gui-Song and Bai, Xiang and Yang, Wen and Yang, Michael Ying and Belongie, Serge and Luo, Jiebo and Datcu, Mihai and Pelillo, Marcello and Zhang, Liangpei},
	journal={IEEE Transactions on Pattern Analysis and Machine Intelligence},
	title={Object Detection in Aerial Images: A Large-Scale Benchmark and Challenges},
	year={2022},
	volume={44},
	number={11},
	pages={7778--7796},
	doi={10.1109/TPAMI.2021.3117983}}

@article{dior,
	title={Object detection in optical remote sensing images: A survey and a new benchmark},
	author={Li, Ke and Wan, Gang and Cheng, Gong and Meng, Liqiu and Han, Junwei},
	journal={ISPRS Journal of Photogrammetry and Remote Sensing},
	volume={159},
	pages={296--307},
	year={2020},
	publisher={Elsevier}
}

@inproceedings{chen2020dynamic,
	title={Dynamic Convolution: Attention over Convolution Kernels},
	author={Chen, Yinpeng and Dai, Xiyang and Liu, Mengchen and Chen, Dongdong and Yuan, Lu and Liu, Zicheng},
	booktitle={Proceedings of the IEEE/CVF Conference on Computer Vision and Pattern Recognition},
	pages={11030--11039},
	year={2020}
}

@inproceedings{dai2017deformable,
	title={Deformable convolutional networks},
	author={Dai, Jifeng and Qi, Haozhi and Xiong, Yuwen and Li, Yi and Zhang, Guodong and Hu, Han and Wei, Yichen},
	booktitle={Proceedings of the IEEE/CVF International Conference on Computer Vision},
	pages={764--773},
	year={2017}
}

@inproceedings{wang2023internimage,
	title={Internimage: Exploring large-scale vision foundation models with deformable convolutions},
	author={Wang, Wenhai and Dai, Jifeng and Chen, Zhe and Huang, Zhenhang and Li, Zhiqi and Zhu, Xizhou and Hu, Xiaowei and Lu, Tong and Lu, Lewei and Li, Hongsheng and others},
	booktitle={Proceedings of the IEEE/CVF Conference on Computer Vision and Pattern Recognition},
	pages={14408--14419},
	year={2023}
}

@inproceedings{wang2018skipnet,
	title={Skipnet: Learning dynamic routing in convolutional networks},
	author={Wang, Xin and Yu, Fisher and Dou, Zi-Yi and Darrell, Trevor and Gonzalez, Joseph E},
	booktitle={Proceedings of the European Conference on Computer Vision},
	pages={409--424},
	year={2018}
}

@article{fan2024not,
	title={Not all layers of llms are necessary during inference},
	author={Fan, Siqi and Jiang, Xin and Li, Xiang and Meng, Xuying and Han, Peng and Shang, Shuo and Sun, Aixin and Wang, Yequan and Wang, Zhongyuan},
	journal={arXiv preprint arXiv:2403.02181},
	year={2024}
}

@ARTICLE{EMO2-DETR,
	author={Hu, Zibo and Gao, Kun and Zhang, Xiaodian and Wang, Junwei and Wang, Hong and Yang, Zhijia and Li, Chenrui and Li, Wei},
	journal={IEEE Transactions on Geoscience and Remote Sensing},
	title={EMO2-{DETR}: Efficient-Matching Oriented Object Detection With Transformers},
	year={2023},
	volume={61},
	pages={1--14},
	doi={10.1109/TGRS.2023.3300154}}

@article{RQFormer,
	title = {RQFormer: Rotated Query Transformer for end-to-end oriented object detection},
	journal = {Expert Systems with Applications},
	volume = {266},
	pages = {126034},
	year = {2025},
	issn = {0957-4174},
	doi = {10.1016/j.eswa.2024.126034},
	author = {Jiaqi Zhao and Zeyu Ding and Yong Zhou and Hancheng Zhu and Wen-Liang Du and Rui Yao and Abdulmotaleb {El Saddik}},
}

@inproceedings{ICLR2025_BEEM,
	author = {Bajpai, Divya Jyoti and Hanawal, Manjesh Kumar},
	booktitle = {International Conference on Learning Representations},
	editor = {Y. Yue and A. Garg and N. Peng and F. Sha and R. Yu},
	pages = {62520--62535},
	title = {BEEM: Boosting Performance of Early Exit DNNs using Multi-Exit Classifiers as Experts},
	volume = {2025},
	year = {2025}
}

@INPROCEEDINGS{roi-transformer,
	author={Ding, Jian and Xue, Nan and Long, Yang and Xia, Gui-Song and Lu, Qikai},
	booktitle={Proceedings of the IEEE/CVF Conference on Computer Vision and Pattern Recognition},
	title={Learning RoI Transformer for Oriented Object Detection in Aerial Images},
	year={2019},
	pages={2844--2853},
	doi={10.1109/CVPR.2019.00296}}

@article{cha2024billion,
	title={A Billion-scale Foundation Model for Remote Sensing Images},
	author={Cha, Keumgang and Seo, Junghoon and Lee, Taekyung},
	journal={IEEE Journal of Selected Topics in Applied Earth Observations and Remote Sensing},
	year={2024},
	publisher={IEEE}
}

@InProceedings{Xie_2021_ICCV,
	author = {Xie, Xingxing and Cheng, Gong and Wang, Jiabao and Yao, Xiwen and Han, Junwei},
	title = {Oriented {R-{CNN}} for Object Detection},
	booktitle = {Proceedings of the IEEE/CVF International Conference on Computer Vision},
	month = {October},
	year = {2021},
	pages = {3520--3529}
}

@inproceedings{li2023lsknet,
	title={Lsknet: Large selective kernel network for remote sensing object detection},
	author={Li, Yuxuan and Hou, Qibin and Zheng, Z},
	booktitle={Proceedings of the IEEE/CVF International Conference on Computer Vision},
	pages={4--6},
	year={2023}
}

@article{zhao2024orientedformer,
	title={OrientedFormer: An end-to-end transformer-based oriented object detector in remote sensing images},
	author={Zhao, Jiaqi and Ding, Zeyu and Zhou, Yong and Zhu, Hancheng and Du, Wen-Liang and Yao, Rui and El Saddik, Abdulmotaleb},
	journal={IEEE Transactions on Geoscience and Remote Sensing},
	year={2024},
	publisher={IEEE}
}

@inproceedings{pu2023adaptive,
	title={Adaptive rotated convolution for rotated object detection},
	author={Pu, Yifan and Wang, Yiru and Xia, Zhuofan and Han, Yizeng and Wang, Yulin and Gan, Weihao and Wang, Zidong and Song, Shiji and Huang, Gao},
	booktitle={Proceedings of the IEEE/CVF Conference on Computer Vision and Pattern Recognition},
	pages={6589--6600},
	year={2023}
}

@inproceedings{yuan2025strip,
	title={Strip R-{CNN}: Large strip convolution for remote sensing object detection},
	author={Yuan, Xinbin and Zheng, ZhaoHui and Li, Yuxuan and Liu, Xialei and Liu, Li and Li, Xiang and Hou, Qibin and Cheng, Ming-Ming},
	booktitle={Proceedings of the AAAI Conference on Artificial Intelligence},
	volume={40},
	number={15},
	pages={12259--12267},
	year={2026}
}

@ARTICLE{10879059,
	author={Zhang, Tong and Zhuang, Yin and Wang, Guanqun and Chen, He and Li, Lianlin and Li, Jun},
	journal={IEEE Transactions on Geoscience and Remote Sensing},
	title={A Unified Remote Sensing Object Detector Based on Fourier Contour Parametric Learning},
	year={2025},
	volume={63},
	pages={1--25},
	doi={10.1109/TGRS.2025.3540085}}

@article{wang2022empirical,
	title={An empirical study of remote sensing pretraining},
	author={Wang, Di and Zhang, Jing and Du, Bo and Xia, Gui-Song and Tao, Dacheng},
	journal={IEEE Transactions on Geoscience and Remote Sensing},
	volume={61},
	year={2022},
	doi={10.1109/TGRS.2022.3181011}
}

@article{wang2022advancing,
	title={Advancing plain vision transformer toward remote sensing foundation model},
	author={Wang, Di and Zhang, Qiming and Xu, Yufei and Zhang, Jing and Du, Bo and Tao, Dacheng and Zhang, Liangpei},
	journal={IEEE Transactions on Geoscience and Remote Sensing},
	volume={61},
	pages={1--15},
	year={2022},
	publisher={IEEE}
}

@article{zeng2024ars,
	title={ARS-{DETR}: Aspect ratio-sensitive detection transformer for aerial oriented object detection},
	author={Zeng, Ying and Chen, Yushi and Yang, Xue and Li, Qingyun and Yan, Junchi},
	journal={IEEE Transactions on Geoscience and Remote Sensing},
	volume={62},
	pages={1--15},
	year={2024},
	publisher={IEEE}
}

@inproceedings{hou2021coordinate,
	title={Coordinate Attention for Efficient Mobile Network Design},
	author={Hou, Qibin and Wang, Changbao and Cheng, Dechao and Cai, Xiaogang and Xu, Gang and Wang, Yuying},
	booktitle={Proceedings of the IEEE/CVF Conference on Computer Vision and Pattern Recognition},
	pages={14310--14320},
	year={2021}
}

@inproceedings{liu2021swin,
	title={Swin transformer: Hierarchical vision transformer using shifted windows},
	author={Liu, Ze and Lin, Yutong and Cao, Yue and Hu, Han and Wei, Yixuan and Zhang, Zheng and Lin, Stephen and Guo, Baining},
	booktitle={Proceedings of the IEEE/CVF Conference on Computer Vision and Pattern Recognition},
	pages={10012--10022},
	year={2021}
}

@inproceedings{swinv2,
	title={Swin transformer v2: Scaling up capacity and resolution},
	author={Liu, Ze and Hu, Han and Lin, Yutong and Yao, Zhuliang and Xie, Zhenda and Wei, Yixuan and Ning, Jia and Cao, Yue and Zhang, Zheng and Dong, Li and others},
	booktitle={Proceedings of the IEEE/CVF Conference on Computer Vision and Pattern Recognition},
	pages={12009--12019},
	year={2022}
}

@article{xu2021vitae,
	title={Vitae: Vision transformer advanced by exploring intrinsic inductive bias},
	author={Xu, Yufei and Zhang, Qiming and Zhang, Jing and Tao, Dacheng},
	journal={Advances in Neural Information Processing Systems},
	volume={34},
	pages={28522--28535},
	year={2021}
}

@article{sun2022fair1m,
	title={{FAIR1M}: A benchmark dataset for fine-grained object recognition in high-resolution remote sensing imagery},
	author={Sun, Xian and Wang, Peijin and Yan, Zhiyuan and Xu, Feng and Wang, Ruiping and Diao, Wenhui and Chen, Jin and Li, Jihao and Feng, Yingchao and Xu, Tao and others},
	journal={ISPRS Journal of Photogrammetry and Remote Sensing},
	volume={184},
	pages={116--130},
	year={2022},
	publisher={Elsevier}
}

@inproceedings{teerapittayanon2016branchynet,
	title={Branchynet: Fast inference via early exiting from deep neural networks},
	author={Teerapittayanon, Surat and McDanel, Bradley and Kung, Hsiang-Tsung},
	booktitle={International Conference on Pattern Recognition},
	pages={2464--2469},
	year={2016},
	organization={IEEE}
}

@inproceedings{uzkent2020learning,
	title={Learning when and where to zoom with deep reinforcement learning},
	author={Uzkent, Burak and Ermon, Stefano},
	booktitle={Proceedings of the IEEE/CVF conference on computer vision and pattern recognition},
	pages={12345--12354},
	year={2020}
}

@article{liu2024scale,
	title={Scale-aware deep reinforcement learning for high resolution remote sensing imagery classification},
	author={Liu, Yinhe and Zhong, Yanfei and Shi, Sunan and Zhang, Liangpei},
	journal={ISPRS Journal of Photogrammetry and Remote Sensing},
	volume={209},
	pages={296--311},
	year={2024},
	publisher={Elsevier}
}

@inproceedings{liu2023seeing,
	title={Seeing beyond the patch: Scale-adaptive semantic segmentation of high-resolution remote sensing imagery based on reinforcement learning},
	author={Liu, Yinhe and Shi, Sunan and Wang, Junjue and Zhong, Yanfei},
	booktitle={Proceedings of the IEEE/CVF International Conference on Computer Vision},
	pages={16868--16878},
	year={2023}
}

@article{mou2021deep,
	title={Deep reinforcement learning for band selection in hyperspectral image classification},
	author={Mou, Lichao and Saha, Sudipan and Hua, Yuansheng and Bovolo, Francesca and Bruzzone, Lorenzo and Zhu, Xiao Xiang},
	journal={IEEE Transactions on Geoscience and Remote Sensing},
	volume={60},
	pages={1--14},
	year={2021},
	publisher={IEEE}
}

@inproceedings{luo2024skipdiff,
	title={Skipdiff: Adaptive skip diffusion model for high-fidelity perceptual image super-resolution},
	author={Luo, Xiaotong and Xie, Yuan and Qu, Yanyun and Fu, Yun},
	booktitle={Proceedings of the AAAI Conference on Artificial Intelligence},
	volume={38},
	number={5},
	pages={4017--4025},
	year={2024}
}

@article{hao2024efficient,
	title={Efficient Adaptive Feature Fusion Network for Remote-Sensing Image Super-Resolution},
	author={Hao, Shuai and Liu, Shuai and Jia, Xu and Lu, Huchuan and He, You},
	journal={IEEE Signal Processing Letters},
	year={2024},
	publisher={IEEE}
}

@inproceedings{chenvision,
	title={Vision Transformer Adapter for Dense Predictions},
	author={Chen, Zhe and Duan, Yuchen and Wang, Wenhai and He, Junjun and Lu, Tong and Dai, Jifeng and Qiao, Yu},
	booktitle={International Conference on Learning Representations},
	year={2023},
}

@article{peng2021conformer,
	title={Conformer: Local Features Coupling Global Representations for Visual Recognition},
	author={Zhiliang Peng and Wei Huang and Shanzhi Gu and Lingxi Xie and Yaowei Wang and Jianbin Jiao and Qixiang Ye},
	journal={arXiv preprint arXiv:2105.03889},
	year={2021},
}

@article{dai2021coatnet,
	title={Coatnet: Marrying convolution and attention for all data sizes},
	author={Dai, Zihang and Liu, Hanxiao and Le, Quoc V and Tan, Mingxing},
	journal={Advances in Neural Information Processing Systems},
	volume={34},
	pages={3965--3977},
	year={2021}
}

@article{lou2025transxnet,
	title={TransXNet: learning both global and local dynamics with a dual dynamic token mixer for visual recognition},
	author={Lou, Meng and Zhang, Shu and Zhou, Hong-Yu and Yang, Sibei and Wu, Chuan and Yu, Yizhou},
	journal={IEEE Transactions on Neural Networks and Learning Systems},
	year={2025},
	publisher={IEEE}
}

@article{li2022next,
	title={Next-{ViT}: Next Generation Vision Transformer for Efficient Deployment in Realistic Industrial Scenarios},
	author={Li, Jiashi and Xia, Xin and Li, Wei and Li, Huixia and Wang, Xing and Xiao, Xuefeng and Wang, Rui and Zheng, Min and Pan, Xin},
	journal={CoRR},
	year={2022}
}

@article{meng2024cta,
	title={CTA-Net: A {CNN}-Transformer Aggregation Network for Improving Multi-Scale Feature Extraction},
	author={Meng, Chunlei and Yang, Jiacheng and Lin, Wei and Liu, Bowen and Zhang, Hongda and Gan, Zhongxue and others},
	journal={arXiv preprint arXiv:2410.11428},
	year={2024}
}

@article{tuia2024artificial,
	title={Artificial Intelligence to Advance Earth Observation: A review of models, recent trends, and pathways forward},
	author={Tuia, Devis and Schindler, Konrad and Demir, Beg{\"u}m and Zhu, Xiao Xiang and Kochupillai, Mrinalini and D{\v{z}}eroski, Sa{\v{s}}o and van Rijn, Jan N and Hoos, Holger H and Del Frate, Fabio and Datcu, Mihai and others},
	journal={IEEE Geoscience and Remote Sensing Magazine},
	year={2024},
	publisher={IEEE}
}

@article{yang2022kfiou,
	title={The KFIoU loss for rotated object detection},
	author={Yang, Xue and Zhou, Yue and Zhang, Gefan and Yang, Jirui and Wang, Wentao and Yan, Junchi and Zhang, Xiaopeng and Tian, Qi},
	journal={arXiv preprint arXiv:2201.12558},
	year={2022}
}

@inproceedings{zheng2018learning,
	title={On Learning Intrinsic Rewards for Policy Gradient Methods},
	author={Zheng, Zeyu and Oh, Junhyuk and Singh, Satinder},
	booktitle={Advances in Neural Information Processing Systems},
	pages={4649--4659},
	year={2018}
}

@article{russakovsky2015imagenet,
	title={Imagenet large scale visual recognition challenge},
	author={Russakovsky, Olga and Deng, Jia and Su, Hao and Krause, Jonathan and Satheesh, Sanjeev and Ma, Sean and Huang, Zhiheng and Karpathy, Andrej and Khosla, Aditya and Bernstein, Michael and others},
	journal={International Journal of Computer Vision},
	volume={115},
	number={3},
	pages={211--252},
	year={2015},
	publisher={Springer}
}

@inproceedings{kool2019stochastic,
	title={Stochastic beams and where to find them: The gumbel-top-k trick for sampling sequences without replacement},
	author={Kool, Wouter and Van Hoof, Herke and Welling, Max},
	booktitle={Proceedings of the International Conference on Machine Learning},
	pages={3499--3508},
	year={2019},
	organization={PMLR}
}

@inproceedings{hu2018senet,
	title={Squeeze-and-Excitation Networks},
	author={Hu, Jie and Shen, Li and Sun, Gang},
	booktitle={Proceedings of the IEEE/CVF Conference on Computer Vision and Pattern Recognition},
	pages={7132--7141},
	year={2018}
}

@INPROCEEDINGS{cai2024pkinet,
	author={Cai, Xinhao and Lai, Qiuxia and Wang, Yuwei and Wang, Wenguan and Sun, Zeren and Yao, Yazhou},
	booktitle={Proceedings of the IEEE/CVF Conference on Computer Vision and Pattern Recognition},
	title={Poly Kernel Inception Network for Remote Sensing Detection},
	year={2024},
	pages={27706--27716},
	doi={10.1109/CVPR52733.2024.02617}}

	\begin{IEEEbiography}[{\includegraphics[width=1in,height=1.25in,clip,keepaspectratio]{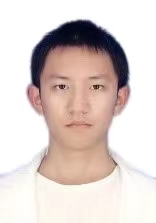}}]{Wenlin Liu}
		received his BS degree from the Nanjing University of Aeronautics and Astronautics in 2021, and he is currently pursuing a doctoral degree at the National University of Defense Technology. His research interests include computer vision, reinforcement learning, and remote sensing object detection.
	\end{IEEEbiography}

	\begin{IEEEbiography}[{\includegraphics[width=1in,height=1.25in,clip,keepaspectratio]{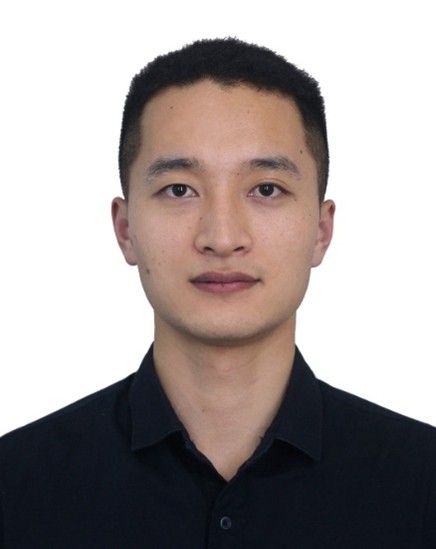}}]{Xikun Hu}
		(Member, IEEE) is an associate professor at College of Electronic Science and Technology, National University of Defense Technology, Changsha, China, and a postdoctoral research fellow at the Department of Geography, The University of Hong Kong, Hong Kong, China. He received a B.S. degree in electronic engineering and an M.S. degree in information and communication engineering from the National University of Defense Technology, Changsha, China, in 2015 and 2017, respectively, and a Ph.D. degree in geoinformatics from the KTH Royal Institute of Technology, Stockholm, Sweden, in 2022. His research interests include multimodal remote sensing image processing and wildfire detection and monitoring.
	\end{IEEEbiography}

	\begin{IEEEbiography}[{\includegraphics[width=1in,height=1.25in,clip,keepaspectratio]{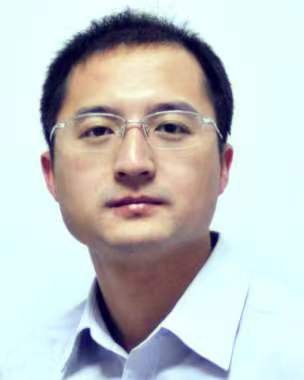}}]{Ping Zhong}
		(Senior Member, IEEE) received the M.S. degree in applied mathematics and the Ph.D. degree in information and communication engineering from the National University of Defense Technology (NUDT), Changsha, China, in 2003 and 2008, respectively.
		
		From 2015 to 2016, he was a Visiting Scholar with the Department of Applied Mathematics and Theoretical Physics, University of Cambridge, Cambridge, U.K. He is currently a Professor with the National Key Laboratory of Science and Technology on ATR, NUDT. He has authored more than 50 peer-reviewed papers in international journals, including TNNLS, TIP, TGRS, JSTSP, and JSTARS. His research interests include computer vision, machine learning, and pattern recognition.
		
		Prof. Zhong received the National Excellent Doctoral Dissertation Award of China in 2011 and was selected for the New Century Excellent Talents in University Program of China in 2013. He currently serves as an Associate Editor for JSTARS.
		
	\end{IEEEbiography}
	\vfill
	
\end{document}